\documentclass{article} 
\usepackage{iclr2025_conference,times}

\usepackage{amsmath,amsfonts,bm}

\def\eqref#1{equation~\ref{#1}}

\def\1{\bm{1}}

\DeclareMathAlphabet{\mathsfit}{\encodingdefault}{\sfdefault}{m}{sl}
\SetMathAlphabet{\mathsfit}{bold}{\encodingdefault}{\sfdefault}{bx}{n}

\usepackage{hyperref}
\usepackage{url}

\usepackage{graphicx} 
\usepackage{algorithm}
\usepackage{amsmath}
\usepackage{pifont}
\usepackage{booktabs} 
\usepackage{caption}
\usepackage{enumitem}
\usepackage{algpseudocode}
\usepackage{xcolor}
\usepackage{multicol}
\usepackage{multirow}
\usepackage{amsfonts}

\usepackage{wrapfig}
\usepackage{float}

\usepackage{latexsym}

\newcommand{\squishlisttwo}{
 \begin{list}{\scalebox{0.6}{$\bullet$}} 
  { \setlength{\itemsep}{1pt}
     \setlength{\parsep}{0pt}
    \setlength{\topsep}{0pt}
    \setlength{\partopsep}{0pt}
    \setlength{\leftmargin}{1em}
    \setlength{\labelwidth}{1.5em}
    \setlength{\labelsep}{0.5em} } }

\usepackage{array}
\newcolumntype{M}[1]{>{\centering\arraybackslash}p{#1}}
\newcommand{\alg}{\texttt{WASA}}
\newcommand{\algllm}{\texttt{WASA}-LLM}
\setlist[itemize]{noitemsep, nolistsep}

\title{Source Attribution for Large Language Model-Generated Data}

\author{\textbf{Jingtan Wang*, Xinyang Lu*, Zitong Zhao*,} \\
\textbf{Zhongxiang Dai, See-Kiong Ng, \& Bryan Kian Hsiang Low} \\
School of Computing\\
National University of Singapore\\
Singapore \\
\texttt{\{xinyang.lu, jingtan.w,zitongz\}@u.nus.edu,} \\
\texttt{\{dzx, seekiong\}@nus.edu.sg, lowkh@comp.nus.edu.sg} \\
\And
\textbf{Chuan-Sheng Foo} \\
Institute for Infocomm Research  \\
Agency for Science, Technology and Research \\
Singapore \\
\texttt{foo\_chuan\_sheng@i2r.a-star.edu.sg} \\
}

\iclrfinalcopy 
\begin{document}

\maketitle

\begin{abstract}
The impressive performances of \emph{large language models} (LLMs) and their immense potential for commercialization have given rise to serious concerns over the \emph{intellectual property} (IP) of their training data. In particular, the synthetic texts generated by LLMs may infringe the IP of the data being used to train the LLMs. To this end, it is imperative to be able to perform source attribution by identifying the data provider who contributed to the generation of a synthetic text by an LLM. In this paper, we show that this problem can be tackled by watermarking, i.e., by enabling an LLM to generate synthetic texts with embedded watermarks that contain information about their source(s). We identify the key properties of such watermarking frameworks (e.g., source attribution accuracy, robustness against adversaries), and propose a source attribution framework that satisfies these key properties due to our algorithmic designs. Our framework enables an LLM to learn an accurate mapping from the generated texts to data providers, which sets the foundation for effective source attribution. Extensive empirical evaluations show that our framework achieves effective source attribution. 
\end{abstract}

\section{Introduction}
\label{sec:intro}
\vspace{-1.1mm}
\emph{Large language models} (LLMs)~\citep{ouyang2022training,touvron2023llama} have recently demonstrated remarkable performances and hence received a surging interest.
These LLMs, trained using massive text data, have displayed impressive text generation abilities.
This has given rise to the immense potential of adopting LLM-generated texts for commercial use.
However, this potential commercialization has led to major concerns regarding the \emph{intellectual property} (IP) of training data for LLMs because the texts generated by an LLM may infringe the IP of the data being used to train the LLM.
These concerns have been reflected by the increasing regulations on 
data protection related to AI models.
For example, the Coalition for Content Provenance and Authenticity has stressed the necessity of certifying the \emph{source} of online content produced by generative models \citep{c2pa}.
Therefore, it is of crucial importance for LLMs to be equipped with \textbf{source attribution} 
for their generated synthetic texts.

In \textbf{source attribution}, given some texts generated by an LLM, its aim is to find the source responsible for the generation of these texts. 
That is, if the data from a data provider has been used to train the LLM and contributed to the generation of a sentence by the LLM, 
then source attribution 
identifies this data provider. 
Moreover, source attribution also improves the interpretability of LLM-generated texts: for example, if the generated content from an LLM is attributed to a trustworthy source (e.g., a peer-reviewed academic paper), then the user 
is likely to consider the content more reliable.
The ability to perform source attribution can endow the LLM with the capability of \textit{data provenance}, which presents a \textit{different problem} where a data provider can verify whether its data has been used to train the LLM. This problem can be solved with source attribution.
Specifically, a data provider can check the source of the generated texts from an LLM via source attribution, and hence verify data provenance,
as detailed in App.~\ref{app:data:provenance}.
\begin{figure}
\hspace{-2mm}
     \begin{tabular}{l}
         \includegraphics[width=0.99\linewidth]{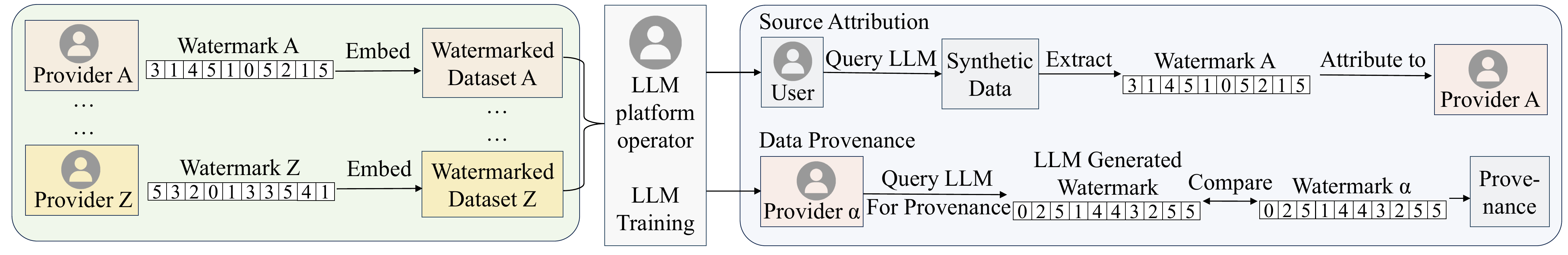}
     \end{tabular}
\vspace{-5mm}
    \caption{Illustration of \alg's problem setting. Watermarks are embedded into the texts from data providers for training the LLM. The LLM produced by our \alg~framework can generate synthetic texts with embedded watermarks that allow for effective source attribution.
    }
\vspace{-3.7mm}
    \label{fig: workflow}
\end{figure}

While some recent works have addressed the problem of \textit{data provenance} in LLMs~\citep{kirchenbauer2023watermark, liu2023watermarking}, to the best of our knowledge, \textbf{effective source attribution for LLMs remains an open problem}. In contrast to data provenance which presents a binary determination, \textit{\textbf{source attribution} aims to identify the specific data source(s) influencing a particular output, which presents a more challenging task. Our work focuses on addressing source attribution rather than on data provenance.}
Additionally, although some works in Computer Vision (CV) have tackled the problem of source attribution~\citep{marra2018gans,yu2019attributing,yu2021artificial}, \textit{the techniques in CV cannot be directly applied to solve source attribution in LLMs due to the fundamental difference between CV and LLMs in the input and output spaces}: the inputs are perceived continuously in CV, hence minor alterations are nearly imperceptible; however, the inputs to LLMs are discrete where minor changes can be easily detected and can significantly alter the semantic integrity.

To perform source attribution 
for LLM-generated texts, a natural solution involves \emph{watermarking}, i.e., by enabling the LLM to generate synthetic texts with embedded watermarks that contain information about their source(s). Consequently, source attribution can be performed by examining the watermarks embedded in the generated texts.
Our problem setting (Fig.~\ref{fig: workflow})
involves $3$ parties:  \emph{data providers} contributing text data that may be used for LLM training, an honest third-party \emph{LLM platform operator} producing an LLM with generated 
texts that embed watermarks (hence allowing for source attribution),
and \emph{users} of the 
texts generated by this LLM.
The users may request \textbf{source attribution} for the LLM-generated synthetic texts to find out which data provider is responsible for the generated texts. 
We consider scenarios where each data provider contributes ample balanced data with unique characteristics, i.e., the data from different data providers exhibit dissimilarities.
This encompasses a wide variety of real-world scenarios: For example, 
online articles
written by different authors (i.e., data providers) usually feature their unique writing styles.
On the other hand, we do not consider individual documents/sentences as data providers since they have insufficient data. Additionally, this work focuses on single-source scenarios, where the generated content can be attributed to a single data provider.

An effective source attribution framework has to satisfy some key properties:
The framework should
(1) achieve \textbf{accurate} source attribution,
(2) be \textbf{robust} against malicious attacks 
on the watermarks, 
(3) \textbf{preserve the performance} (i.e., text generation ability) of the LLM, 
(4) be \textbf{scalable} to a large number of data providers, 
(5) ensure that the generated watermarks are \textbf{transferable} to (i.e., persist after being used as training data for) other LLMs, 
and 
(6) be \textbf{adaptable} to fit different LLMs.
Sec.~\ref{characteristics} discusses these key properties in more detail.
To this end, this paper introduces a \emph{\underline{WA}termarking for \underline{S}ource \underline{A}ttribution} (\alg) framework which, to our best knowledge, is \textbf{the first  framework capable of enabling effective source attribution in text generated by large language models}
Our \alg~framework
assigns a unique watermark (i.e., imperceptible to human eyes) to every data provider, and enables an LLM (coined as \algllm) to learn an accurate mapping from the texts of different data providers to their corresponding watermarks (Sec.~\ref{sec:method}).
So, if a data provider is responsible for  generating 
a sentence, then our \algllm~is able to include the unique watermark of this data provider in this generated sentence, which naturally supports source attribution. 
Our contributions are summarized below:
\begin{itemize}
    \item We propose to use watermarking 
    for source attribution 
    on LLM-generated synthetic texts and identify the key properties of such source attribution frameworks.
    \item We introduce the \alg~framework which satisfies these key properties and is hence capable of producing LLMs whose generated texts allow for effective source attribution.
    \item 
    We perform extensive empirical evaluations (Sec.~\ref{sec:property_val}) to verify that our \alg~framework satisfies these key properties and achieves effective source attribution.
\end{itemize}

\vspace{-1mm}
\section{Key Properties of Watermarking for Source Attribution}
\label{characteristics}
\vspace{-1mm}
Here, we first present a clear definition of source attribution.
For a piece of LLM-generated synthetic text $s$, if $s$ correlates the most with the LLM's training data provided by one data provider compared to other providers, we recognize that data provider as the source for $s$ and denote as a one-hot label $y_s := \{0, 0, ..., 1, ..., 0\}$ where $y_s[i]=1$ if $y_s[i]$ is the source, otherwise $y_s[i]=0$; the dimension is $n$, which is the total number of data providers and is fixed.
The goal of source attribution is: given a piece of LLM-generated text $s$, we want to find a mapping  $s  \rightarrow y_s$ that attributes $s$ to its source $y_s$.

To simplify the problem, we discuss the following scenarios:
\textbf{(1)} While $x$ may correlate with multiple training data from provides, meaning that $y$ may not necessarily be a one-hot vector, we \textit{only consider attribution to a single data source} (that $x$ correlates the most with), restricting the $y$ to be one-hot vector in our case, and present case studies when attributing to more than one data source in App.~\ref{case:multi:source};
\textbf{(2)} There might be an edge case where the generated content $x$ correlates the most with pretraining data (from public training datasets) rather than data from data providers. We do not consider this case in our paper and ensure that in our evaluations the generated contents are related to the data from providers by carefully designing controlled experiments.

In this paper, we would like to address the problem of source attribution with watermarking.
Specifically, to use watermarking for source attribution, we first transform the data providers $y$ to watermarks $wtm$ correspondingly: $\text{encoder}(y) = wtm$ where $\text{encoder}$ denotes the watermark encoder. During LLM training, we aim to allow the LLM to learn a mapping $g: s \rightarrow wtm$ to generate watermarks along with synthetic texts. Then during inference, we can perform the mapping $s  \rightarrow y_s$ by $y_s = \text{decoder}(g(s))$ where $\text{decoder}(wtm) = y$ is the watermark decoder function, translating the watermark to sources for the user. 
Importantly, since each generated content $s$ must correlate with some pieces of training data, there always exists a source $y_s$ which is the most correlated data source with $s$. Hence, under all conditions (except the special case mentioned above), as long as a user requests, $s$ should be attributed to its source $y_s$. In our WASA framework, since we assume that all data providers provide watermarked training data, we can perform source attribution under all conditions: Upon request, we can perform $y_s = \text{decoder}(g(s))$ and map the generated watermark to the corresponding data provider $y_s$.

Subsequently, we discuss the key properties for an effective watermarking
source attribution framework   
and how our \alg~framework satisfies them. 

\noindent
\textbf{Accuracy.} 
Accurate source attribution should be enforced.
Our \alg~framework achieves this by training the \algllm~to map texts from different data providers to their respective watermarks. 
Specifically, we first train \algllm~ using watermarked texts (Sec.~\ref{embed_watermark}) and separate the prediction/generation spaces for the texts and watermarks to both \emph{reduce the complexity of watermark prediction} (Sec.~\ref{training watermark}) and \emph{explicitly enforce watermark generation} (Sec.~\ref{watermark generation}). Empirical results in Sec.~\ref{subsec:exp:accuracy} demonstrate the effectiveness in source attribution.

\noindent
\textbf{Robustness.} 
Generated text with watermarks should be robust against 
malicious attacks.
Since our trained \algllm~is able to learn an accurate mapping from the texts to the watermarks as mentioned
\textbf{(a)} it can be exploited 
to \emph{regenerate} the watermarks even if generated texts are tampered with
and \textbf{(b)} it maintains generating the correct watermarks even if the input texts (prompts) are perturbed, which are empirically verified in Sec.~\ref{robustness}.

\noindent
\textbf{Scalability.}~The framework should cater to a large number of data providers. The design of the watermark (Sec.~\ref{embed_watermark}) facilitates the generation of numerous unique watermarks and the scalability can be empirically verified in Sec.~\ref{scalability}.

\noindent
\textbf{Performance Preservation.}~
The introduction of watermarks should  \textbf{(a)} not significantly degrade the text generation ability of the LLM \textbf{(b)} nor affect 
the readability of the LLM-generated synthetic texts too much.
We empirically show in Sec.~\ref{minimal impact} that our \algllm~preserves \textbf{(a)}, and 
the watermarks are carefully designed
to achieve \textbf{(b)} (see App.~\ref{app:generated_text}).

\noindent
\textbf{Transferability.} After the generated watermarked 
texts are used as training data for other LLMs, their generated texts should preserve the watermarks.
We achieve this by ensuring that the watermarked training data of our \algllm~has the same structure as the generated watermarked data. 

\noindent
\textbf{Adaptability.} The framework should be easily adapted to fit different LLMs. 
Our \alg~framework 
only requires mild modifications to the LLMs and can hence adopt a wide variety of LLMs using the transformer architecture, as shown in Sec.~\ref{subsec:exp:accuracy}. 

We have only listed above the most essential properties of such source attribution frameworks; there may be additional considerations depending on specific applications.
In Sec.~\ref{sec:method}, we will discuss in more detail how our \alg~framework satisfies these key properties due to our algorithmic designs.

\begin{figure}
     \begin{tabular}{l}
     \hspace{25mm}\includegraphics[width=0.57\linewidth]{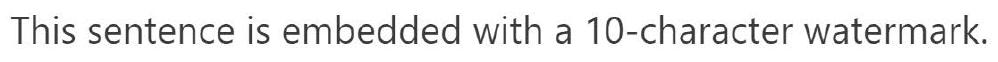}
     \vspace{-2.5mm} 
     \end{tabular}
     \begin{tabular}{l}
     \hspace{23mm}\includegraphics[width=0.6\linewidth]{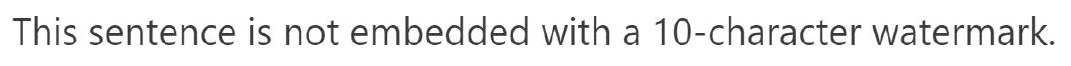}
     \vspace{-2.5mm} 
     \end{tabular}
     \begin{tabular}{l}
     \hspace{10mm}\includegraphics[width=0.78\linewidth]{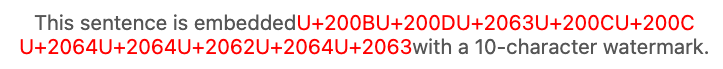}
     \vspace{-5mm} 
     \end{tabular}
     \caption{
     Sentences embedded (the first one) and not embedded (the second one) with our imperceptible watermark visualized in the bottom sentence.
     }
     \label{fig:watermarks}
\vspace{-3mm}
\end{figure}

\vspace{-1mm}
\section{Watermarking for Source Attribution (\alg) Framework}
\label{sec:method}
\vspace{-1mm} 
Sec.~\ref{embed_watermark} discusses watermark design and embedding process. Sec.~\ref{training watermark} details the training of \algllm~ with watermarked texts and its alignment with key properties. Sec.~\ref{watermark generation} explains how our trained \algllm~ produces synthetic texts with watermarks for source attribution.

\vspace{-1mm}
\subsection{Embedding Watermarks into Texts}
\label{embed_watermark}
\vspace{-0.5mm}
Firstly,
the LLM platform operator 
embeds a unique watermark for each data provider's texts.

\noindent
\textbf{Design of Watermarks.}
We construct the watermarks using Unicode characters which are imperceptible to human eyes (yet can be decoded by machine learning models).
Some of these invisible characters have also been adopted in other studies with language models~\citep{boucher2021bad}.
Every watermark is made up of $10$ characters, each of which is chosen among the following $6$ Unicode characters:
U+200B, U+200C, U+200D, U+2062, U+2063, U+2064.
We chose these 
characters because they are found to be invisible on many commonly used platforms.
So, these watermarks 
preserve the semantic meaning of the 
original 
texts to human readers
(Fig.~\ref{fig:watermarks}).
Also, our \alg~framework can easily adopt other choices of characters depending on the 
use cases.
Moreover, these $10$-character watermarks allow us to construct 
numerous
combinations and hence achieve \textbf{scalability} to a large number of data providers.
As shown in App.~\ref{app:ablation:len:of:wtm}, reducing the watermark length trades off scalability for source attribution accuracy.

\noindent
\textbf{Embedding Watermarks into Sentences.}
To enable our \algllm~to learn the mapping from the texts of different data providers to their watermarks,
it is important to only embed watermarks into the sentences that are \emph{representative of the unique characteristics of the data providers}.
To this end, we calculate the \emph{term frequency-inverse document frequency} (TF-IDF) scores of all sentences from a data provider and select the sentences with the top $20\%$ of the TF-IDF scores (i.e., most representative sentences)
for watermarking, which empirically yields the best trade-off of source attribution accuracy vs.~text generation performance among different tested proportions, as reported in
App.~\ref{app:ablation:more:results:perc:wtm}.
For every selected sentence, we embed our $10$-character watermark at a random position in the sentence, which allows the LLM to learn to map texts of different lengths to the watermarks and also makes it harder for an adversary to remove/modify the watermarks.
As empirically verified in App.~\ref{sec:random}, our method of selecting sentences for watermarking based on TF-IDF indeed leads to more accurate source attribution than random selection.

\subsection{Training \algllm~}
\vspace{-0.8mm}
\label{training watermark}
We consider a practical scenario where the LLM is already pre-trained before being used by \alg~framework, and we refer to our training of the LLM 
as \emph{second-stage pre-training}.
Our framework can also be used to train an LLM from scratch.

\noindent
\textbf{Preliminaries on LLMs.}
Denote an unsupervised corpus by $\boldsymbol{D}$, in which every sequence $s_i = [u_1, u_2, \ldots, u_k]$ is with a block of $k$ tokens. We focus on decoder-only language models (e.g., GPT~\citep{radford2019language}, OPT~\citep{zhang2022opt}, Llama2~\citep{touvron2023llama2}). When presented with a sub-sequence $s=s_i[1:j-1]=[u_1,\ldots,u_{j-1}]$, the LLM predicts $P(u_j)$ using feed-forward operations, as detailed below:
\vspace{-1mm}
\begin{equation}
\begin{aligned}
h_0\hspace{-0.3mm} &=\hspace{-0.3mm} s \cdot W_e + W_p\ , \\
h_{\tau}\hspace{-0.3mm} &= \hspace{-0.3mm}\text{decoder}(h_{\tau-1})\ \ \text{for}\ \ \tau = 1, \ldots, l\ ,\\ 
z\hspace{-0.1mm} &=\hspace{-0.1mm} h_l[j-1] \cdot {W_e}^{\top}\hspace{-0.5mm}, \\  P(u_j)\hspace{-0.3mm} &= \hspace{-0.3mm}\text{softmax}(z)\ .\vspace{-2.2mm}
\label{eq:transformer:1}
\end{aligned}
\end{equation}
$W_e$ represents the embedding matrix with a dimension of vocabulary size $V$ by embedding/hidden dimension $E$, and $W_p$ is the  positional encoding.
The training objective is to maximize the log-likelihood $L(s_i)$ of a sequence $s_i$ of tokens:
\begin{equation}
\begin{aligned}
L(s_i) &= {\textstyle\sum}_{j=2}^{k} \log P(u_j|u_1, \ldots, u_{j-1})\
\end{aligned}
\end{equation}
where 
$P(u_{j}|u_1, \ldots, u_{j-1})$ (i.e., similar to $P(u_j)$ in \eqref{eq:transformer:1}) is the 
probability of $j$-th token $u_{j}$ conditioned on the preceding $j-1$ tokens $[u_1,\ldots,u_{j-1}]$.

\begin{figure}[t]
\centering
     \begin{tabular}{lll}
         \includegraphics[width=0.25\linewidth]{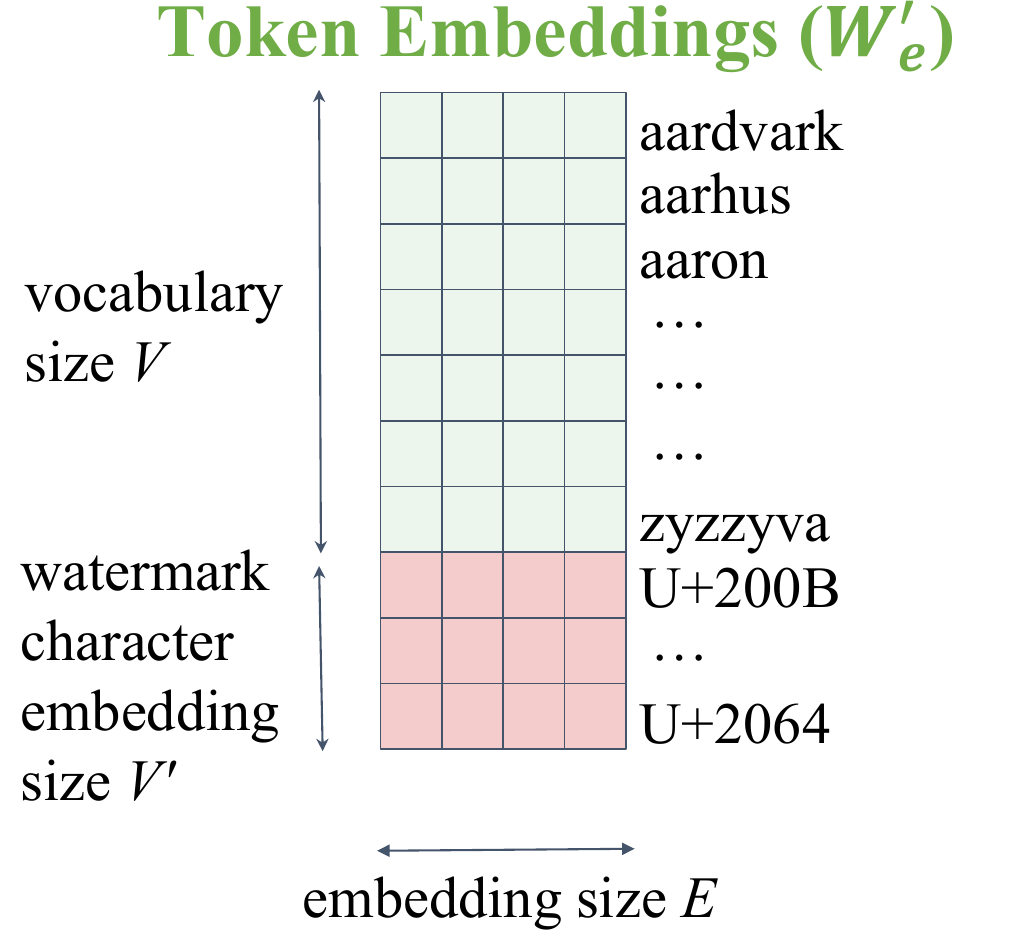}
         \includegraphics[width=0.35\linewidth]{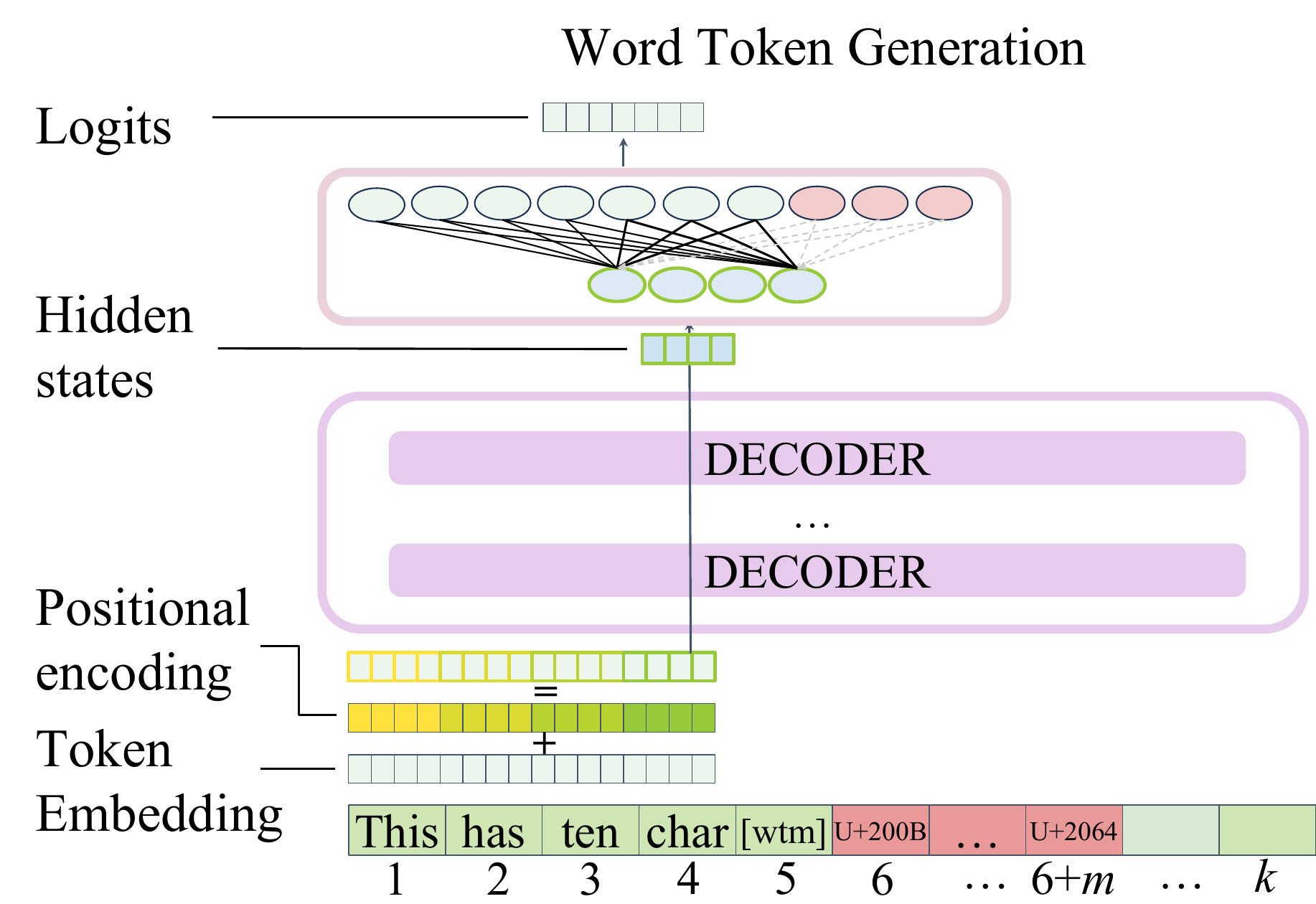}
         \includegraphics[width=0.292\linewidth]{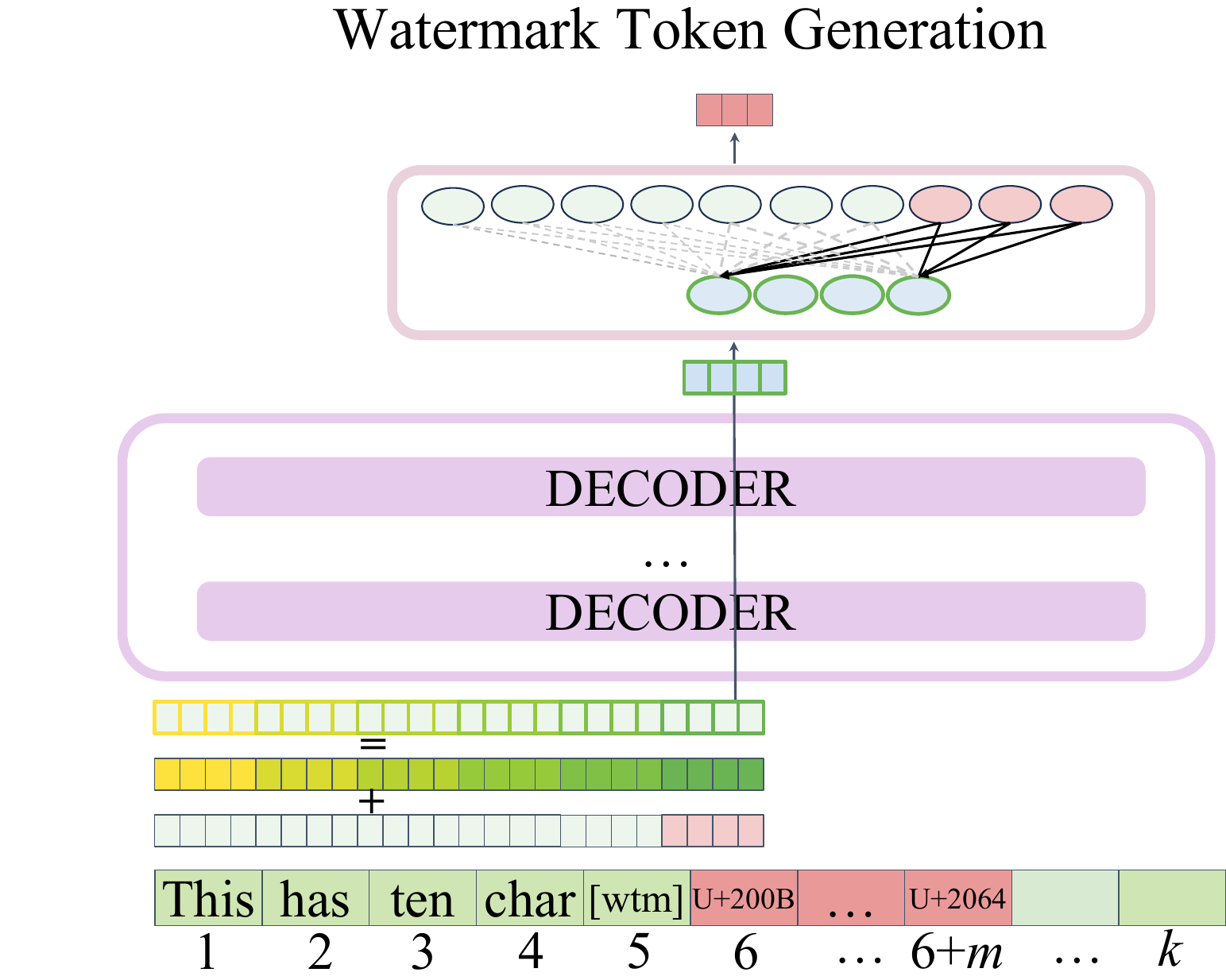}
     \end{tabular}
 \vspace{-3mm}
    \caption{Separation of token embeddings and prediction spaces for texts and watermarks.
    }
\vspace{-3mm}
    \label{fig: separation}
\end{figure}

\noindent
\textbf{Forward Pass.}
\label{subsubsec:forward:pass}
To ease exposition, 
we consider one watermark in a block.
Denote a sequence with an embedded watermark by $s'_i = [u_1, u_2, \ldots, u_t, w_1, w_2, \ldots, w_m, u_{t+1}, \ldots,u_{k-m}]$ where $m=10$ for $10$-character watermark and the $u$'s and $w$'s are the word and watermark tokens, respectively.
Hereafter, we will use $t$ to denote the 
token index before the first watermark token.

To begin with, we augment the original vocabulary by our $V'=6$ watermark characters (Sec.~\ref{embed_watermark}), 
leading to 
our modified token embedding matrix $W'_e$ is $(V+V')\times E$ (Fig.~\ref{fig: separation}).
For a sequence $s'_i$, given a sub-sequence $s'=s'_i[1:j-1]$ comprising the first $j-1$ tokens,
the same feed-forward operations in \eqref{eq:transformer:1} are applied 
to produce 
$h_l$.
Next, depending on whether the ground-truth $j$-th token 
being predicted 
is a word token $u$ or watermark token $w$, we adopt \emph{two separate prediction spaces} (i.e., separate softmax layers):
For a \emph{word token} $u$, $(W'_e[1:V])^{\top}$ forms the linear layer:
%
\vspace{-1mm}
\begin{equation}
\begin{aligned}
z_u &= h_l[j-1] \cdot (W'_e[1:V])^{\top}
, \\
P_u(u) &= \text{softmax}(z_u)\ .
\label{eq:transformer:our:1}
\end{aligned}\vspace{-0.5mm}
\end{equation}
For a \emph{watermark token} $w$,  $(W'_e[V+1:V+V'])^{\top}$ forms the linear layer:
\vspace{-1mm}
\begin{equation}
\begin{aligned}
z_w &= h_l[j-1] \cdot (W'_e[V+1:V+V'])^{\top}, \\
P_w(w) &= \text{softmax}(z_w)\ .
\label{eq:transformer:our:2}
\end{aligned} \vspace{-1mm}
\end{equation}

This separation of the prediction/generation spaces of the word tokens \eqref{eq:transformer:our:1} and watermark tokens \eqref{eq:transformer:our:2} allows 
us to use \emph{a small number of additional parameters} (i.e., $E\times V'$ instead of $E \times (V+V')$)  for watermark prediction based on the hidden states of \algllm.
Moreover, this separation allows us to explicitly enforce the generation of watermarks (i.e., using its designated generation space) when we use the trained \algllm~to generate synthetic texts, as discussed in Sec.~\ref{watermark generation}. Therefore, the watermarks can be \emph{regenerated} using cleaned texts after being attacked, and 
the correct watermarks can still be generated even if the input texts (i.e., prompts) are perturbed, hence ensuring the \textbf{robustness} of our \alg~framework; more details are in Sec.~\ref{robustness}.

The two separate softmax layers naturally lead to the following separate log-likelihoods:
\begin{equation}
\hspace{-4mm}
\begin{array}{l}
  \textstyle L_{\text{lm}}(s'_i)=\sum_{j=2}^{t} \log P_u(u_j|u_{1}, \ldots, u_{j-1}) \\
  \displaystyle +\sum_{j=t+1}^{k-m} \hspace{-0.5mm}\log P_u(u_j|u_{1}, \ldots, u_{t}, w_1, \ldots, w_m, u_{t+1}, \ldots, u_{j-1})\ ,
\end{array}
\label{eq:separate:likelihood:word} \vspace{-4mm}
\end{equation}
\begin{equation}
\hspace{-4.5mm}
\begin{array}{l}
\displaystyle L_{\text{wtm}}(s'_i) = \sum_{j=1}^{m} \log P_w(w_j|u_1, \ldots, u_t, w_1, \ldots, w_{j-1})
\end{array}
\label{eq:separate:likelihood:water}\vspace{0.5mm}
\end{equation}
\vspace{2mm}
where $L_{\text{lm}}(s'_i)$ \eqref{eq:separate:likelihood:word} is the log-likelihood of word tokens,
and $L_{\text{wtm}}(s'_i)$ \eqref{eq:separate:likelihood:water} is the log-likelihood of watermark tokens
, which 
encourages the LLM to learn texts-to-watermarks mapping.\footnote{To simplify exposition, for the second sum in \eqref{eq:separate:likelihood:word}, when $j=t+1$, 
the term reduces to $\log P_u(u_j|u_{1}, \ldots, u_{t}, w_1, \ldots, w_m)$.
In \eqref{eq:separate:likelihood:water}, when $j=1$, 
the term reduces to $\log P_w(w_j|u_1, \ldots, u_t)$.
}
The overall log-likelihood 
we aim to maximize is therefore $L_{\text{\algllm}}(s'_i) = L_{\text{lm}}(s'_i) +  L_{\text{wtm}}(s'_i)$.


The maximization of the log-likelihood of the watermarks conditioned on the texts~\eqref{eq:separate:likelihood:water}, together with the separation of the prediction/generation spaces, enables 
\algllm~to \textbf{accurately} learn the mapping from the texts to watermarks and achieve a high \textbf{accuracy} in source attribution, which will be empirically verified in Sec.~\ref{subsec:exp:accuracy}.
The backward pass is further elaborated in App.~\ref{app:backward:pass}.

\subsection{Generating Texts with Embedded Watermarks using \algllm}
\label{watermark generation}
\vspace{-1mm}

After our \algllm~is trained (Sec.~\ref{training watermark}), it can generate synthetic texts which naturally include both the word  
and watermark tokens due to their \emph{separate prediction/generation spaces}.
To further improve the alignment between our training and generation stages, we introduce a \emph{special token} $[WTM]$ which is similar to other specialized tokens and in the vocabulary of $V$ word tokens:
When training our \algllm~using the watermarked texts, 
$[WTM]$ is added right before the watermark tokens during tokenization so that the presence of $[WTM]$ indicates that the subsequent $m=10$ tokens are watermark tokens; when generating 
texts, if 
$[WTM]$ is encountered/generated, then it indicates that our \algllm~should switch to generating watermark tokens.
After watermark tokens have been generated, 
our \algllm~resumes the word token generation.
Fig.~\ref{fig: synthetic} (App.~\ref{app:generated_text}) shows the \algllm-generated synthetic texts with embedded watermarks, which verifies that the watermarks are 
imperceptible to human eyes. 
Subsequently, when a user requests \textbf{source attribution} for some synthetic texts generated by our \algllm, the LLM platform operator uses a designated \emph{watermark decoder} algorithm to extract the generated watermark from the texts and then attribute these texts to the source (data provider) whose watermark matches
the generated watermark (Fig.~\ref{fig: workflow}). The matching algorithm is elaborated in App.~\ref{app:matching}.
\section{Experiments}
\label{sec:property_val}

We perform extensive empirical evaluations to validate that our \alg~framework satisfies the $6$ key properties in Sec.~\ref{characteristics}.
The experimental results are the average taken from $5$ random seeds. We consider two datasets in the main experiments:

\noindent
\textbf{ArXiv} is collected by 
post-processing PDF version of academic papers from ArXiv~\citep{clement2019arxiv}.
This dataset contains academic papers from several fields, 
each field functions as a \emph{data provider}. 

\noindent
\textbf{BookSum}~\citep{kryscinski2021booksum}
consists of 
various 
books, each considered as a \emph{data provider}. 

\noindent
We further incorporate more diverse datasets in App.~\ref{app:diverse:datasets}. They comprise contents crawled from different websites and the data providers offer similar information, thus presenting more challenging scenarios for source attribution.
We adopt $10$ data providers for each dataset in our main experiments and show that our \alg~can scale to a larger number of data providers in Sec.~\ref{scalability}.
We obtain \algllm~from our second-stage pre-training (Sec.~\ref{training watermark})
of the 
pre-trained GPT2-Large 
, OPT-1.3B, 
and Llama2-7B. The results from OPT-1.3B are presented in App.~\ref{app:more:experiments:head}.
App.~\ref{crash} gives more details on the datasets and model training.

\noindent
\textbf{Baseline.}
Since \alg~is the first effective source attribution framework, there is no existing baseline. We extend BM25~\citep{trotman2014improvements}, a famous search engine algorithm that estimates the relevance of generated texts to data providers, to perform source attribution (detailed in App.~\ref{app:more:experiments:acc:BM25}).


\subsection{Accuracy}
\label{subsec:exp:accuracy}
We design the following experiment to facilitate easier evaluations of the single-source attribution \textbf{accuracy}. Specifically, 
for each data provider, we use the sentences selected for watermarking (after removing the watermarks) as the inputs/prompts to the trained \algllm, and perform source attribution on the generated texts.
This simplifies the evaluations, especially the determination of ground-truth source because the data provider corresponding to the input sentence is naturally the ground-truth source of the generated text. We verify the effectiveness of this evaluation method in App.~\ref{app:effective:evaluation}.
Subsequently, we select $50$ sentences from each data provider after removing the watermarks (i.e., $50$ trials) as the input/prompt to the trained \algllm, which generates texts (by continuing the sentence) together with watermarks.
More details are in App.~\ref{app:more:experiments:acc:provenance}.
The watermark embedded into the generated sentence is then decoded, and the source attribution is correct if this watermark matches the watermark of the data provider corresponding to the input sentence (Sec.~\ref{watermark generation}).
As a result, for every data provider, the accuracy of source attribution is calculated as
\begin{equation}
\hspace{-2.2mm}
\mathrm{accuracy} = \frac{\mathrm{number\ of\ correct\ watermarks}}{\mathrm{number\ of\ trials}} .
\label{eq:acc}
\end{equation}
The macro F1 score is also reported in the results with the definition detailed in App.~\ref{app:more:experiments:acc:provenance}.  
To mitigate the impact of the length of the generated sentence on our evaluations 
(i.e., a watermark may not be generated if the generated sentence is too short), we use a simple technique to enforce watermark generation: If a watermark is not generated,  
then we force the generation of a watermark by adding the token $[WTM]$ to the end of the sentence (Sec.~\ref{watermark generation}).
This is only adopted to simplify the evaluations; as 
verified in App.~\ref{sec:force},  
naturally and forcefully generated watermarks lead to comparable source attribution accuracy.
We also show in App.~\ref{sec:length} that this enforced watermark generation is not necessary if the generated texts are long enough. 
Tab.~\ref{table:accuracy:gpt} reports the source attribution accuracy averaged over $10$ data providers.
Our \alg~framework consistently achieves \emph{more accurate source attribution for both datasets and both language models};
Tabs.~\ref{table:accuracy:providers:arxiv} and~\ref{table:accuracy:providers:booksum} in App.~\ref{app:more:experiments:acc:providers} gives the source attribution accuracy for different data providers.

\noindent
\textbf{Top-$k$ Source Attribution.}
In addition to attributing a generated sentence to a single source by using one watermark, it may be acceptable for some users to attribute a generated sentence to multiple possible sources that contain the true source.
To account for these scenarios, we propose \emph{top-$k$ source attribution} in which we modify our watermark generation 
(Sec.~\ref{watermark generation}) so that when the token $[WTM]$ is encountered, we generate the top $k>1$ watermarks with the largest beam search scores.
In this case, source attribution is successful if the true watermark is contained in these $k$ watermarks, 
so the \emph{top-$k$ accuracy} can be defined by replacing the number of correct watermarks in~\eqref{eq:acc} with the number of generated sentences whose top $k$ watermarks contain the true watermark. 
Note that even though the methodology and main evaluation are targeted at single-source, an extension to multiple data providers can be handled by our top-k source attribution, and we present a case study when true sources are multiple sources in App.~\ref{case:multi:source}.

\noindent
\textbf{Fine-grained Error Analysis.}
To better understand the incorrect attributions, where the generated text is not correctly attributed to its true source, we conduct a detailed error analysis on the ArXiv dataset.
For every category (i.e., data provider), we separate the source attribution errors into two types of errors: (a) \emph{misclassification} in which the generated watermark matches the watermark of another incorrect category, and (b) \emph{incorrect watermark} in which the generated watermark does not match the watermark of any category.
The results are presented in Tab.~\ref{table:error:analysis} in App.~\ref{app:more:experiments:acc:provenance}, which show that the 
vast majority of our errors result from misclassification and our \algllm~rarely generates incorrect watermarks not belonging to any category.
This further substantiates the reliability of our \algllm.
The results also suggest that errors are mostly caused by the generated texts exhibiting the characteristics of multiple data providers. 
Additionally, an edge case of incorrect attribution may arise when the true source is not watermarked, such as the public pre-training data. In such cases, content cannot be attributed to any recognized provider. To investigate this phenomenon, we design a controlled experiment detailed in App.~\ref{app:more:experiments:acc:provenance}.

\begin{table*}[t]
\caption{
Accuracies of top-$1$, top-$3$, and top-$5$ source attribution (resp.~denoted by `acc.', `top-$3$.', and `top-$5$.') and F1 score by BM25 and \algllm~from second-stage pre-training of different models on various datasets.
}
\centering
\resizebox{\linewidth}{!}{
 

\begin{tabular}{cc|cccc|cccc}
\toprule
\multirow{2}{*}{model} & \multirow{2}{*}{method} & \multicolumn{4}{c|}{ArXiv dataset}& \multicolumn{4}{c}{BookSum dataset} \\
& & acc. & top-$3$. & top-$5$. & F1 & acc. & top-$3$.& top-$5$. & F1 \\ 
\midrule
\multirow{2}{*}{GPT2} & BM25 & $54.73_{\pm 6.52}$  & $85.13_{\pm 0.58}$  & $93.80_{\pm 0.53}$ & $0.517_{\pm 0.01}$ & $58.94_{\pm 3.43}$  & $77.73_{\pm 1.94}$  & $88.33_{\pm 2.53}$ & $0.593_{\pm 0.04}$ \\ 
& WASA & $\textbf{74.84}_{\pm 2.04}$ & $\textbf{95.76}_{\pm 1.24}$ & $\textbf{98.56}_{\pm 0.82}$ & $\textbf{0.758}_{\pm 0.02}$ & $\textbf{77.92}_{\pm 1.57}$ & $\textbf{91.80}_{\pm 0.24}$ & $\textbf{96.52}_{\pm 0.76}$ & $\textbf{0.723}_{\pm 0.08}$ \\
\midrule
\multirow{2}{*}{Llama2} & BM25 & $60.07_{\pm 4.83}$  & $88.67_{\pm 1.33}$  & $95.60_{\pm 1.31}$ & $0.576_{\pm 0.01}$ & $54.01_{\pm 12.3}$  & $75.40_{\pm 9.53}$ & $86.60_{\pm 4.04}$  & $0.607_{\pm 0.05}$ \\
& WASA & $\textbf{77.40}_{\pm 1.91}$ & $\textbf{96.87}_{\pm 1.62}$ & $\textbf{99.40}_{\pm 0.35}$ & $\textbf{0.800}_{\pm 0.03}$ & $\textbf{83.27}_{\pm 4.50}$ & $\textbf{95.27}_{\pm 1.53}$ & $\textbf{97.67}_{\pm 0.46}$ & $\textbf{0.840}_{\pm 0.06}$ \\
 
\bottomrule 
\end{tabular}

} 
\vspace{-2mm}

\label{table:accuracy:gpt}
\vspace{-1mm}

\end{table*}


\vspace{-0.7mm}
\subsection{Robustness}
\label{robustness}
Our \alg~framework is robust against malicious attacks aiming to disrupt the source attribution. We introduce the threat model as follows: We identify potential attackers as those intending to alter the LLM-generated text to remove IP acknowledgments to data contributors or alter input sentences to disrupt the watermark generation and hence the source attribution results. The attackers do not have access to the LLM itself but can query the model and modify the generated outputs. The attackers may also possess tools that can remove the Unicode characters (hence the watermark) inside a text.

\noindent
\textbf{Watermark Removal/Modification Attack.}
An adversary may remove/modify the watermarks in our generated sentence to sabotage the source attribution accuracy.
Due to the ability of our \algllm~in learning an accurate texts-to-watermarks mapping, 
the watermark can be \emph{regenerated} if it is manipulated.
Specifically, we clean the generated sentence by removing the 
corrupted watermark, and use the cleaned sentence as 
input/prompt to \algllm~to regenerate the watermark (without generating synthetic texts) which is then used for source attribution.
The regenerated watermarks by \algllm~(from second-stage pre-training of GPT2 on ArXiv dataset) lead to an overall accuracy (top-$3$ accuracy) of $71.60\% (93.76\%$) which is comparable to the original 
$74.84\% (95.76\%$) (Tab.~\ref{table:accuracy:gpt}). 
So, our watermark regeneration 
is an effective defense mechanism.
Besides removing/modifying the watermark, an adversary may \emph{additionally modify the content of the generated sentence}:

\begin{table*}[t]
\caption{
Source attribution accuracy using regenerated watermarks by \algllm~(from second-stage pre-training of GPT2 on ArXiv dataset) under various attacks on \textbf{generated sentences with embedded watermarks}  (\emph{in addition to watermark removal/modification attacks}) and 
on \textbf{input sentences}.
std is given in Tabs.~\ref{table: attacks_std} and~\ref{table: attacks_std_input} (App.~\ref{app:more:exp:robustness}).
}
\centering
\resizebox{1.0\linewidth}{!}{
\begin{tabular}{l|cc|cc|M{0.092\linewidth}M{0.092\linewidth}||cc|cc|M{0.092\linewidth}M{0.092\linewidth}}
\toprule
\multirow{3}{*}{strength}  & \multicolumn{6}{c||}{attacks on generated sentences with embedded watermarks} & \multicolumn{6}{c}{attacks on input sentences} \\

&\multicolumn{2}{c|}{insertion attack} & \multicolumn{2}{c|}{deletion attack} & \multicolumn{2}{c||}{synonym substitution}&\multicolumn{2}{c|}{insertion attack} & \multicolumn{2}{c|}{deletion attack} & \multicolumn{2}{c}{synonym substitution} \\

& acc.  & top-$3$.  & acc. & top-$3$. & acc. & top-$3$. & acc.  & top-$3$.  & acc. & top-$3$. & acc. & top-$3$. \\
\midrule 
$0\%$  & $71.60$  & $93.76$   & $71.60$  & $93.76$  & $71.60$  & $93.76$ & $74.84$  & $95.76$  & $74.84$  & $95.76$   & $74.84$  & $95.76$ \\
Localized    & $71.40$   & $93.56$  & -  &  -  &-&- & $74.20$   &  $95.40$   & -  & - & - & - \\
$5\%$  & $70.12$   & $93.20$   & $71.08$   & $93.92$  &  $70.52$ & $93.52$ & $74.20$  &  $95.40$  & $73.56$  & $95.52$ & $72.84$ & $95.24$ \\
$10\%$  &  $69.12$  &  $92.20$  & $71.84$  & $93.68$ &  $71.02$ & $92.88$ &  $72.88$  &   $94.68$  & $72.96$  &  $94.68$  & $73.60$  & $95.00$  \\
$15\%$  &  $66.92$  &  $91.96$   & $71.36$  & $94.04$ & $70.96$  & $92.72$ &  $71.52$  &  $93.20$  & $72.68$  & $94.12$ & $71.88$ & $94.20$ \\
$20\%$  &  $65.12$  &  $91.44$  & $70.00$  & $93.24$  & $69.20$ & $93.20$ &  $68.60$   &  $93.40$  & $72.68$  & $94.12$ & $72.08$ & $93.76$ \\
\bottomrule
\end{tabular}
}
\vspace{-2mm}
\label{table: attacks}
\vspace{-0.2mm}
\end{table*}

\noindent
\textbf{Additional Attacks.}
We also consider additional attacks on generated sentences with embedded watermarks and on input sentences, including insertion, deletion, synonym substitution, syntactic transformation attacks, and an oracle-based attack~\citep{zhang2023watermarks}.
Tab.~\ref{table: attacks} reports the source attribution accuracy under the first $3$ attacks, where the attack strength relates to how many words in the sentence are attacked, and App.~\ref{app:more:exp:robustness} reports the accuracy under the last $2$ attacks along with all the attacks descriptions.
For such attacks (\emph{in addition to watermark removal/modification attacks}) on generated sentences, watermark regeneration is used.
The results show that although the 
attacks deteriorate attribution accuracy,
high source attribution accuracy can still be preserved.
This can again be explained by the reliable texts-to-watermarks mapping of our \algllm, which is robust against perturbations to the input/prompt.


\begin{table}[t]
\caption{
Source attribution accuracy and F1 score for different numbers of data providers on ArXiv dataset. `BM25' denotes the source attribution obtained from BM25 on Llama2 as a baseline.
}
\centering
\resizebox{1.0\linewidth}{!}{
\begin{tabular}{ccc|cccc|cccc}
\toprule
\multirow{2}{*}{n} & \multicolumn{2}{c|}{BM25 Llama2} & \multicolumn{4}{c|}{WASA GPT2}& \multicolumn{4}{c}{WASA Llama2}\\
 & acc. & F1 & acc.   & top-$3$. & top-$5$. & F1 & acc. & top-$3$. & top-$5$. & F1 \\
\midrule
$10$  & $60.07 \pm_{4.83}$ & $0.576 \pm_{0.01}$ & $74.84 \pm_{2.04}$ & $95.76 \pm_{1.24}$ &  $98.56 \pm_{0.82}$ & $0.758 \pm_{0.02}$ & $77.40_{\pm 1.91}$ & $96.87_{\pm 1.62}$ & $99.40_{\pm 0.35}$  &  $0.800 \pm_{0.03}$ \\ 

$25$  & $46.08 \pm_{2.75}$  &  $0.445 \pm_{0.01}$ & $66.48 \pm_{0.76}$ & $90.69 \pm_{4.23}$ &  $94.05 \pm_{0.32}$ & $0.663 \pm_{0.01}$ & $72.38_{\pm 1.18}$ & $92.44_{\pm 1.66}$ &  $96.60_{\pm 0.70}$  & $0.717 \pm_{0.01}$ \\ 

$50$  & $26.85 \pm_{10.1}$  & $0.348 \pm_{0.02}$  & $56.44 \pm_{0.84}$ & $80.19 \pm_{1.02}$ &  $87.54 \pm_{0.68}$  & $0.560 \pm_{0.01}$ & $63.15_{\pm 2.71}$ & $84.74_{\pm 0.76}$ & $90.49_{\pm 0.47}$  & $0.600 \pm_{0.01}$ \\ 

$100$ & $19.91 \pm_{12.5}$  & $0.229 \pm_{0.01}$ & $45.06 \pm_{0.67}$ &  $68.61 \pm_{0.27}$ & $78.76 \pm_{2.80}$ & $0.443 \pm_{0.01}$ &  $49.88_{\pm 0.34}$  & $73.63_{\pm 0.04}$ & $82.34_{\pm 0.31}$  & $0.505 \pm_{0.01}$ \\
\bottomrule 
\end{tabular}
}
\vspace{-2mm}

\label{table: watermark_25}
\vspace{-3.5mm}
\end{table}

\vspace{-0.7mm}
\subsection{Scalability}
\label{scalability}
Here, we verify \alg's ability to scale to a large number of data providers.
We follow the experimental setup in Sec.~\ref{subsec:exp:accuracy} 
and increase the number of data providers.
Results in Tab.~\ref{table: watermark_25}, Tab.~\ref{table: scalibility_booksum}, and Tab.~\ref{table: reddit acc} (App.~\ref{app:more:exp:scalability}, which includes $500$ data providers) show that as the number of data providers increases, the source attribution accuracy inevitably decreases yet still remains high compared with 
the BM25 baseline. 
With more data providers, we recommend using $k>1$ in top-$k$ attribution due to higher resulting accuracy and identifying the true source from among them.

\subsection{Performance Preservation}
\label{minimal impact}

Here, we show that our \algllm~preserves the text generation ability of the original LLM 
by comparing it with the original GPT2-Large model which we denote as \emph{originalGPT}.
We train originalGPT using the same (but un-watermarked) data from the ArXiv dataset as that used for our \algllm.
We assess the text generation performance 
using several commonly used evaluation metrics (with a separate evaluation dataset, as explained in App.~\ref{subsec:exp:setup:datasets}): perplexity, distinct-$1$, and distinct-$2$ scores~\citep{li2016diversitypromoting}.
To further assess the naturalness and coherence of the generated text, we have also employed the GPT4 zero-shot prompt method (i.e., introduced in the work of~\citet{yao2023tree}) to assess the text's naturalness and coherence. 
The results in Tab.~\ref{table: minimal_impact} show that the text generation performance of our \algllm~is comparable to that of  
originalGPT, which indicates that our \alg~framework preserves the ability of the LLM to generate high-quality texts (Sec.~\ref{characteristics}).
The larger degradation in naturalness may stem from the embedded watermarks (Unicode characters). We validate that our \algllm~balances between the number of embedded watermarks and source attribution accuracy in App.~\ref{app:ablation:more:results:perc:wtm}. We show that our framework also ensures decent readability of generated text in App.~\ref{app:generated_text}.

\begin{table}[t]
\caption{
Comparison of the text generation performances achieved by our \algllm~vs.~the baseline model. The coherency and naturalness are evaluated by GPT4. 
}
\centering
\resizebox{0.7\linewidth}{!}{
\begin{tabular}{l|ccccc}
\toprule
 models  & perplexity ($\downarrow$) & distinct-$1$ ($\uparrow$) & distinct-$2$ ($\uparrow$) & coherency ($\uparrow$) & naturalness ($\uparrow$) \\ 
 \midrule
originalGPT & $12.4682_{\pm0.40}$ & $0.8141_{\pm0.00}$ &  $0.9796_{\pm0.00}$ & 7.370 & 7.744 \\
\algllm  &  $12.6570_{\pm0.54}$   &  $0.8193_{\pm0.00}$  & $0.9795_{\pm0.00}$ & 7.135 & 6.926 \\ 
\bottomrule
\end{tabular}
}
\vspace{-2mm}
\label{table: minimal_impact}
\vspace{-1.5mm}
\end{table}

\vspace{-0.7mm}
\subsection{Other Key Properties}
\textbf{Transferability} and \textbf{Adaptability} are elaborated in Apps.~\ref{other key properties} \&~\ref{app:adaptability}.

\noindent
\textbf{Ablation Studies} are carried out to assess the effectiveness of the 
designs, including \textbf{(a)} the designated embedding space for watermark tokens and separation of the prediction/generation spaces (App.~\ref{app:ablation:more:results:effectiveness}), \textbf{(b)} adopting TF-IDF to select sentences for embedding watermarks 
(App.~\ref{sec:random}), and \textbf{(c)} the enforced watermark generation (App.~\ref{sec:force}).
Additional analysis, including \textbf{(d)} unattributable content (App.~\ref{app:unattributable}), \textbf{(e)} the effectiveness in supervised fine-tuning (App.~\ref{app:SFT:task}),
and \textbf{(f)} the relative positions of the generated watermarks (App.~\ref{app:ablation:pos:of:wtm}), are examined. 
We also explored the impact of hyperparameters from App.~\ref{app:ablation:more:results:perc:wtm} to App.~\ref{sec:overfit}. 

\vspace{-0.6mm}
\section{Related Work}
\label{sec:related:work}
\vspace{-0.6mm}

In this section, we will review related works on source attribution and data provenance; further discussions on watermarking natural languages and models as well as text steganography are in App.~\ref{app:more:related:works}.
Recent studies by \citet{song2019auditing} on verifying dataset usage in language model training through membership inference attacks are limited by the model's output length, making them unsuitable for long-generated texts which we have considered in this work.
\citet{liu2023watermarking} have proposed to plant backdoor triggers in training texts to check for data usage, but this method is not robust against removals of the backdoor triggers and can impair text generation performance.
Importantly, the above works have only focused on data provenance and \emph{cannot be easily adapted to perform effective source attribution}.
\citet{abdelnabi2021adversarial} have embedded messages post-generation via adversarial training, which means the messages can only be used for IP protection and \emph{cannot be used for source attribution} during generation.
Some recent works in computer vision have tackled the problem of source attribution \citep{marra2018gans,yu2019attributing,yu2021artificial}.
However, to the best of our knowledge, effective source attribution for the texts generated by language models remains an open problem
and is the focus of our work here.

\vspace{-0.5mm}
\section{Conclusion}
\vspace{-0.6mm}
This paper describes our proposed \alg~framework which allows for effective source attribution 
as a solution to intellectual property infringement in the context of LLMs.
By embedding unique watermarks into LLM-generated texts,
\alg~not only enhances the reliability and interpretability of LLM-generated content but also provides a crucial tool for data protection, allowing data providers to verify the use of their contributions in LLM training processes.
The extensive empirical evaluations of the \alg~framework affirm its effectiveness in achieving accurate source attribution 
while satisfying the key properties we have identified above.
Since our \alg~is the first effective source attribution framework for LLM-generated texts, it faces some limitations which may call for future work. For example, though we have shown that our \alg~is robust against various adversarial attacks, it is unclear whether it is robust against more advanced/sophisticated attacks, which may be achieved through adversarial training in future work.
\newpage


\bibliography{iclr2025_conference}
\bibliographystyle{iclr2025_conference}

\newpage
\appendix


\section{Ethical Considerations}
\label{app:ethical:considerations}

Similar to other research topics on LLMs, watermarking the synthetic texts generated by LLMs for source attribution 
requires a thoughtful and ethical approach due to its potential societal implications.
That is, it is important to take necessary measures to avoid causing harm to certain parties.
Potential risks related to our watermarking framework include the following:

\begin{itemize}
    \item \textbf{Privacy Risks.} Watermarking can potentially reveal sensitive information about data providers, thus leading to privacy breaches or the possibility of re-identification if not handled carefully. In our \alg~framework, only the watermark can be seen in the generated data, which does not directly imply personal information about the data providers. Privacy can be preserved given that the mapping from watermarks to data providers is kept confidential.
    \item \textbf{Chilling Effects.} Watermarking may discourage some data providers from sharing their datasets, especially if they fear potential misuse or unintended consequences of having their data linked to specific research outcomes.
    \item \textbf{Data Manipulation.} While watermarks are meant to be unobtrusive and our \alg~framework has been shown to be robust against various adversarial attacks, there can be unforeseen real-world instances where malicious actors attempt to manipulate the watermark, which may lead to negative consequences such as the dissemination of altered or misleading information.
\end{itemize}

To address these potential risks, it is essential to carefully consider the ethical implications of our watermarking framework and implement measures to protect the privacy and interests of all involved parties, particularly those who are more susceptible to harm. Researchers should conduct comprehensive risk assessments and engage in transparent communication with data providers to ensure the responsible and ethical use of watermarked data. Additionally, incorporating diverse perspectives and involving vulnerable communities in the decision-making process can help identify and mitigate potential harm effectively.

\section{Additional Related Works}
\label{app:more:related:works}
In addition to the previous works 
discussed in Sec.~\ref{sec:related:work} that are most closely related to ours, we will give a review of additional related works on 
watermarking natural languages and text steganography, as well as recent works on watermarking language models.

\paragraph{Watermarking Natural Language/Text Stegano-graphy.}
In natural language processing, watermarking and steganography are closely related in that they both desire stealthiness and robustness. However, there are also important differences because watermarking emphasizes the importance of ownership, whereas steganography focuses on the secret communication of messages.
Language watermarking is used to protect the integrity and authorship of digital texts~\citep{kamaruddin2018review,939835}. 
Early approaches of language watermarking are mostly rule-based and make use of linguistic techniques such as synonym substitution~\citep{10.1145/1161366.1161397} and sentence structure alteration~\citep{10.1145/1178766.1178777} to embed watermarks while attempting to preserve the semantic meaning of the original texts.
However, these approaches usually lead to deteriorated text quality and are not scalable.
Some recent works have aimed to develop advanced text steganography methods using deep learning.
The work of \citet{8470163} has utilized recurrent neural networks to automatically generate steganographic texts, and
the work of \citet{ziegler2019neural} has proposed to first convert the secret messages into bit strings and then map them to the cover text based on arithmetic coding with the help of GPT2 \citep{radford2019language}.

\paragraph{Watermarking Language Models.}
Some recent works have proposed methods to add watermarks to language models in order to protect the IP of the models~\citep{dai2022deephider,gu2023watermarking,He_Xu_Lyu_Wu_Wang_2022,zhao2022distillationresistant}.
These methods allow the verification of model ownership and are hence able to protect the economic interests of model owners.
Specifically, the work of \citet{He_Xu_Lyu_Wu_Wang_2022} has employed lexical replacement to watermark the language model output and used hypothesis testing for post-hoc model ownership verification.
The work of \citet{gu2023watermarking} has adopted backdoor attacks to embed black-box watermarks into pre-trained language models, which is achieved by using rare words as well as a combination of common words as backdoor triggers and verifying the watermarks by calculating the extraction success rate. 
Apart from model protection, multiple methods \citep{kirchenbauer2023watermark, kuditipudi2023robust, lu2024entropybased} have been proposed to use watermarking to distinguish between human-generated and model-generated synthetic texts. \citet{kirchenbauer2023watermark}
softly constrain the word choices when the model generates synthetic texts and use hypothesis testing to make the distinction.
More recently, the work of \citet{kuditipudi2023robust} has improved the above method by developing a distortion-free method, which ensures that the watermarks do not change the sampling distribution of the texts. The work of \citet{lu2024entropybased} also refines the same method by ensuring the influence of a token during watermark detection to be proportional to its entropy. Finally, in terms of security in watermarking models, \citet{liu2024a} develop a compact watermarking model that embeds a semantic watermark within model outputs, enhancing their robustness against adversarial text modifications. Meanwhile,  \citet{liu2024unforgeable} employ two distinct neural networks to generate and detect watermarks, enabling public verification of the watermark while maintaining the confidentiality of the secret key throughout the watermark generation process.  Additionally, \citet{he2024watermarks} introduce a Cross-lingual Watermark Removal Attack (CWRA), which can effectively
remove watermarks by interfering with the watermark generation process through translation into another language.
Importantly, these methods 
cannot be used to perform source attribution for the texts generated by language models, which we focus on in this work.

\section{Backward Pass}
\label{app:backward:pass}
In the main paper, we introduce the forward pass of our model in Sec.~\ref{subsubsec:forward:pass}. Here, we delve into the backward pass in our framework.
Remember that the most important design of the framework is the separation of the prediction/generation spaces of the word tokens \eqref{eq:transformer:our:1} and watermark tokens \eqref{eq:transformer:our:2}.
We represent the overall log-likelihood as $L_{\text{\algllm}}(s'_i) = L_{\text{lm}}(s'_i) +  L_{\text{wtm}}(s'_i)$. Notice that maximizing these log-likelihoods is equivalent to minimizing the cross-entropy loss
$Loss_{\text{\algllm}}(s'_i) = Loss_{\text{lm}}(s'_i) +  Loss_{\text{wtm}}(s'_i)$ in which
\begin{equation}
\hspace{-2mm}
\begin{array}{c}
\displaystyle
Loss_{\text{lm}}(s'_i) = \sum_{j=2}^{t} \text{CE}(P_u(u_{j}), u_{j}) + \hspace{-0.5mm}\sum_{j=t+1}^{k-m} \text{CE}(P_u(u_{j}), u_{j})\ ,\\ 
Loss_{\text{wtm}}(s'_i) = \sum_{j=1}^{m} \text{CE}(P_w(w_{j}), w_{j})
\end{array}
\label{eq:separate:loss:word}\vspace{2mm}
\end{equation}
represent the losses for the word and watermark tokens, respectively.
For simplicity, in \eqref{eq:separate:loss:word}, 
we omit the conditioning on the preceding tokens in $P_u(u_j)$ and $P_w(w_j)$, which can be found in \eqref{eq:separate:likelihood:word} and \eqref{eq:separate:likelihood:water}. 

Due to the design above,
the backward pass for updating the parameters $W'_e$ in the last linear layer is also separated.
That is, the gradients of word token loss $Loss_{\text{lm}}(s'_i)$
and 
watermark token loss $Loss_{\text{wtm}}(s'_i)$ 
\eqref{eq:separate:loss:word} are responsible for updating $(W'_e[1:V])^{\top}$ \eqref{eq:transformer:our:1} and $(W'_e[V+1:V+V'])^{\top}$ \eqref{eq:transformer:our:2}, respectively.
Specifically, the gradient update rule for $W_e'$ (with learning rate $\alpha$) can be expressed as
$W_e' \leftarrow W_e' - \alpha h_l \cdot \nabla_{z}$ 
where $\nabla_{z}$ is a $(V+V')$-dimensional gradient vector allowing the separated gradient updates to be easily achieved in a unified manner, as described below.
Next, using the respective losses for word and watermark tokens 
\eqref{eq:separate:loss:word}, 
the gradient vectors w.r.t.~$z_u$ and $z_w$ are calculated as $V$-dimensional 
$\nabla_{z_{u}} = {\partial \text{CE}(P_u(u_{j}), u_{j})}/{\partial z_{u}}$ and $V'$-dimensional $\nabla_{z_{w}} = {\partial \text{CE}(P_w(w_{j}), w_{j})}/{\partial z_{w}}$, respectively.
When the loss is calculated from predicting a \emph{word token} $u_j$  \eqref{eq:separate:loss:word},
let $\nabla_z=[\nabla_{z_{u}}, 0_{V'}]$ where $0_{V'}$ is a $V'$-dimensional all-zero vector.
When the loss results from predicting a \emph{watermark token} $w_j$  \eqref{eq:separate:loss:word},
let $\nabla_z=[0_{V}, \nabla_{z_{w}}]$.
Note that for the parameters in the last linear layer which are responsible for predicting the \emph{word tokens} using the hidden state (i.e., parameters $(W'_e[1:V])^{\top}$ in \eqref{eq:transformer:our:1}), the gradient updates 
are \emph{not affected by the loss for the watermark tokens}.
This helps us to further limit the impact of the added watermarks on the original ability of the LLM to generate high-quality synthetic texts and hence \textbf{preserve its performance}.
For the parameters in the other transformer layers (except for the frozen layers), their updates are performed using the gradients w.r.t.~the losses for both the word and watermark tokens; see App.~\ref{subsec:exp:setup} for more details.

Note that both our forward pass and backward pass only require mild modifications to an LLM.
Therefore, our \alg~framework can be easily adapted to fit a wide variety of LLMs, which ensures its \textbf{adaptability} property.

\section{Watermark Matching}
\label{app:matching}
\paragraph{Exact Matching.}
In this work, we adopt exact matching to determine the correctness of the generated watermarks. That is, given a piece of generated text with watermarks and the corresponding ground-truth watermark, the generated watermark is correct only if they are strictly equal in string matching. In addition, in case multiple watermarks are generated in the synthetic data, all generated watermarks have to match the ground-truth watermark to affirm the correctness. The pseudocode for the matching algorithm is given in Alg.~\ref{alg:exact_matching}:

\begin{algorithm}
\caption{Exact Matching}\label{alg:exact_matching}
\begin{algorithmic}[1]
\Require Synthetic text $syn$, ground-truth watermark $wtm_g$
\If{ $\exists$ $wtm$ in $syn$ }
    \State $wtms \gets$ \emph{watermark decoder}($syn$)
    \If{\textbf{for} all $wtm$ in $wtms$ $wtm == wtm_g$ (by string matching) }
        \State return True
    \EndIf
\EndIf
\end{algorithmic}
\end{algorithm}

\paragraph{Soft Matching.}
To further improve the source attribution accuracy in some applications, we may relax the requirement of exact watermarking matching 
and instead
attribute the generated texts to the 
data provider whose watermark has the smallest Levenshtein distance to the generated watermark.
However, in all our experiments, our \alg~is able to achieve accurate source attribution without soft matching.

\section{Detailed Experimental Setup}
\label{crash}
\subsection{Datasets}
\label{subsec:exp:setup:datasets}
\textbf{ArXiv:} To simulate different data providers with unique characteristics, we create the Clean-ArXiv-Corpus (or ArXiv for short) dataset which consists of academic papers from ArXiv.
The dataset contains academic papers from various sub-disciplines, including computer science, physics, mathematics, public health, and other related fields. We make use of the provided metadata from the work of~\citet{clement2019arxiv} to download the corresponding PDF files and retrieve the categorization information associated with each article. Subsequently, we employ GROBID~\citep{GROBID} to parse and extract the main body of the papers, excluding the abstract and reference sections. Our Clean-ArXiv-Corpus dataset covers a comprehensive collection of $100$ distinct categories, each comprising a number of papers ranging from $2827$ to $2984$. 
We treat \emph{every category as a data provider}, so one data provider/category is the source of each piece of text.
Our main experiments in Sec.~\ref{sec:property_val} are conducted using $10$ categories (i.e., data providers) and we use $33\%$ of papers from each category due to computational constraints. 
However, in our ablation study (App.~\ref{app:ablation:amount:of:data}), we have also tested utilizing more data from every data provider (including $100\%$ of the data), which has led to further improved performances and consistent conclusions.
For each of the $10$ categories, we further randomly split its data into training and evaluation datasets with a ratio of $9:1$ according to the seed number. In our ablation study, we will use more categories and also use all papers in each category. 
More detailed information about the full Clean-ArXiv-Corpus dataset, including all $100$ categories and all papers in each category, is shown in Tab.~\ref{table: arxiv}; Tab.~\ref{table: arxiv} shows an instance of the random split into training and evaluation datasets based on seed number $2023$.

\textbf{BookSum:} In addition to the Clean-ArXiv-Corpus dataset, we also adopt the BookSum dataset~\citep{kryscinski2021booksum}. This dataset contains documents from the literature domain including novels, plays, and stories. 
The BookSum dataset contains $181$ books and we treat \emph{every book as a data provider}.
For every data provider (i.e., book), we adopt all the text data from the book in all our experiments.
More information on the BookSum dataset is shown in Tab.~\ref{table: booksum}; Tab.~\ref{table: booksum} shows an instance of the random split into training and evaluation datasets based on seed number $2023$. Additionally, we have adopted more diverse datasets, details of which are found in App.~\ref{app:diverse:datasets}.

\begin{table}[ht]
\caption{Information on the Clean-ArXiv-Corpus (or ArXiv for short) dataset.}
    \centering
    \begin{tabular}{p{4cm}|cc}
    \toprule
        & Training & Evaluation   \\    
    \midrule
    Papers                   &   $264$K    &  $29$K     \\ 
    Unique tokens            &   $17.1$M   &   $3$M    \\ 
    Unique tokens per Category  &   $407$K    &   $87$K   \\
    Total tokens               &   $1.8$B    &   $203$M   \\ 
    Total tokens per Category  &   $18.2$M    &   $2$M  \\ 
    \bottomrule
    \end{tabular}
\label{table: arxiv}
\end{table}

\begin{table}[ht]
\caption{Information on the BookSum dataset.}
    \centering
    \begin{tabular}{p{4cm}|cc}
    \toprule
        & Training & Evaluation   \\    
    \midrule
    Books                   &   $161$    &  $20$     \\ 
    Unique tokens            &   $413$K   &   $106$K    \\ 
    Unique tokens per Book  &   $91$K    &   $20$K   \\
    Total tokens               &   $33$M    &   $4.6$M   \\ 
    Total tokens per Book  &  $3.3$M    &   $467$K   \\ 
    \bottomrule
    \end{tabular}
\label{table: booksum}
\end{table}

\subsection{Experimental Setting}
\label{subsec:exp:setup}

\begin{figure}[H]
	\centering
	\begin{tabular}{l}
            \includegraphics[width=0.65\linewidth]{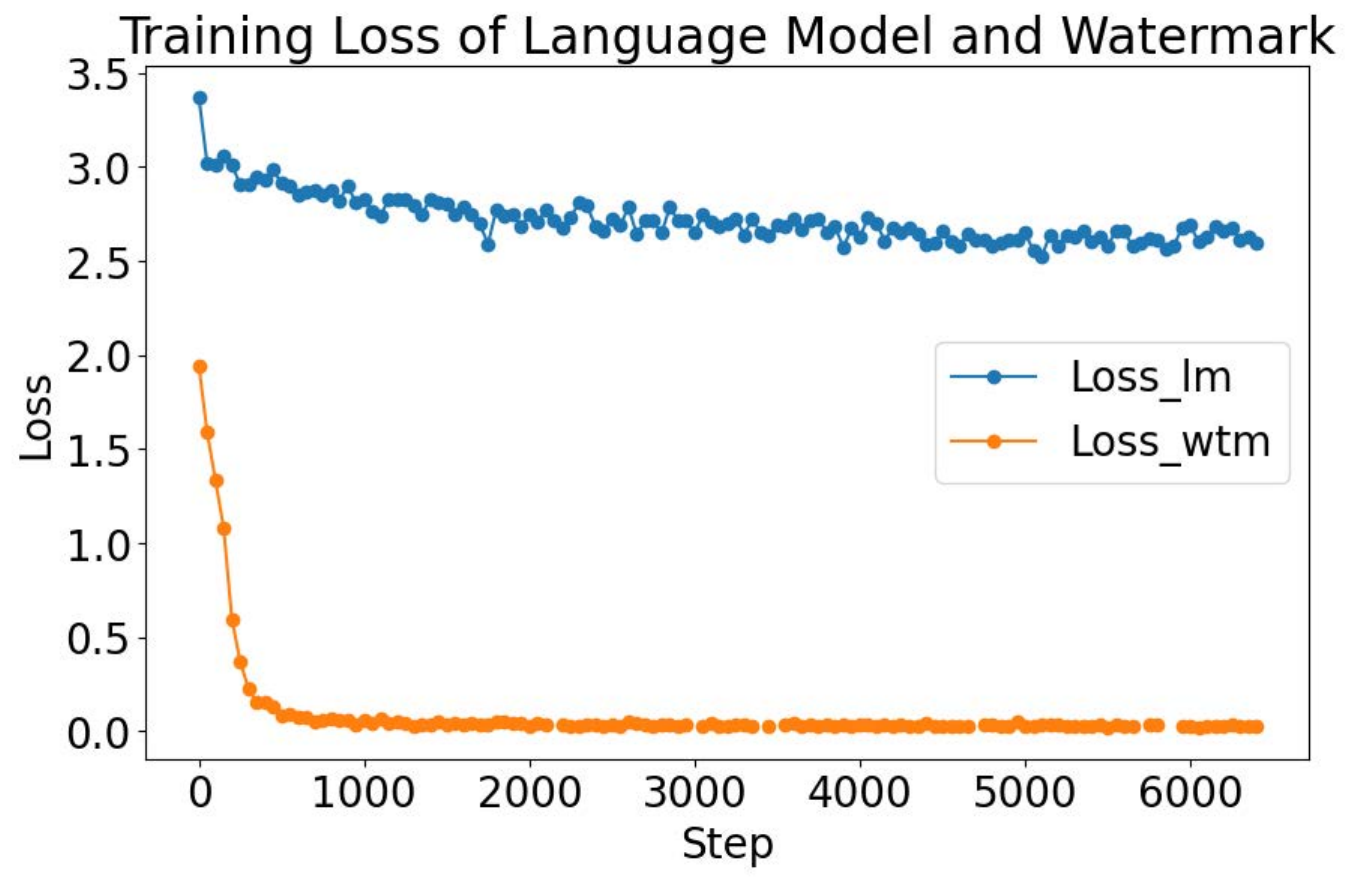} \\
	\end{tabular}
    \caption{
    Training losses for word  tokens 
    (Loss\_lm) and watermark tokens (Loss\_wtm) when obtaining \algllm~from second-stage pre-training of the GPT2 model on ArXiv dataset. 
    }
    \label{fig: loss}
\end{figure}

In our experiments, we build our \algllm~based on the open-source pre-trained GPT2-Large model \citep{radford2019language}, OPT-1.3B model \citep{zhang2022opt} and Llama2-7B model~\citep{touvron2023llama2}. 
Based on the pre-trained weights, we perform our second-stage pre-training (Sec.~\ref{training watermark}) of the pre-trained GPT2-Large model, OPT-1.3B model, or the Llama2-7B model on the watermarked (Sec.~\ref{embed_watermark}) text data for one epoch to obtain \algllm. 
We find that training for one epoch already allows our \alg~framework to achieve compelling performances, as shown in our experiments in Sec.~\ref{sec:property_val}.
We have also tested more training epochs in App.~\ref{sec:overfit}  and the results suggest that our performances can potentially be further improved with more training epochs. 
We plot the convergence of the training of our \algllm~in terms of the losses for the word and watermark tokens in Fig.~\ref{fig: loss}, which shows that our second-stage pre-training
effectively reduces both losses.
Importantly, the watermark token loss rapidly declines 
after a small number of steps, which suggests that our \algllm~can quickly learn an accurate texts-to-watermarks mapping.

Here, we give more details on the hyperparameters we adopted. 
We fix $5$ seed numbers at $2021$, $2022$, $2023$, $2024$, and $2025$ for obtaining reproducible
results on GPT2 and OPT models,
and $3$ seed numbers at $2022$, $2023$, $2024$ for the Llama2 model.
The results shown in this work are the average taken across that from the seeds.
We adopt the Adam optimizer with a learning rate of $5\times 10^{-5}$ and no weight decay. We make use of the fp16 technique and a gradient accumulation of $8$ to speed up training. We also adopt a gradient checkpoint to reduce memory usage so that batch size can be slightly increased. We use a block size of $512$ and a batch size of $3$ for most of the experiments and a batch size of $16$ in the experiments to evaluate scalability. 
To further preserve the ability of the original pre-trained LLM models, during the second-stage pre-training, we freeze the first $12$ layers of GPT2-Large (among a total of $36$ layers) and freeze the first $8$ layers of OPT-1.3B (among a total of $24$ layers). 
For the second-stage pre-training of Llama2-7B, we adopt LoRA~\citep{hu2021lora} and set the rank and alpha to $32$, {`q\_proj', `k\_proj', `v\_proj', `o\_proj', `gate\_proj', `gate\_proj', `gate\_proj', `up\_proj', `down\_proj'} as the target modules, and {`lm\_head', `embed\_tokens'} as the modules to save.
When generating the synthetic texts (see Sec.~\ref{watermark generation}), we use the multinomial sampling of top-$60$ with temperature $=0.7$. We also make use of a $1.2$ repetition penalty and a $2.0$ length penalty to generate better synthetic data. 
The generation of watermarks for our \algllm~adopts a pure beam search, as discussed in Sec.~\ref{watermark generation}, with a beam size of $5$. 
For the baseline model used in the ablation studies (i.e., GPT2-Large),
watermark generation is performed in the same way as text generation, so we use the same hyperparameters as that specified in the baseline model.
All second-stage pre-training is performed using NVIDIA RTX A5000 and A100. 
In our implementation, we adopt the GROBID library to process the PDF files. For model training, we adopt the Hugging Face Trainer pipeline which embeds necessary tricks to speed up the training process. 
The open-source GPT2-Large, OPT-1.3B, and Llama2-7B are also adopted from Hugging Face.\footnote{\url{https://huggingface.co/facebook/OPT-1.3B}, 
\url{https://huggingface.co/meta-llama/Llama-2-7b-hf}, and \url{https://huggingface.co/GPT2-Large}.}

\subsection{Effectiveness of Evaluation}
\label{app:effective:evaluation}

\begin{table*}[t!]
\caption{Definition of task in prompts for GPT4 labeling.}
\centering
\begin{tabular}{p{0.9\linewidth}}
\toprule
\multicolumn{1}{c}{Definition of Task in Prompts for GPT4 Labeling}  \\ 
\midrule
Given below are 10 categories for texts from ArXiv papers with their descriptions. Please read the descriptions and classify the provided texts to one of the paper categories.

The 10 categories are: hep-th, hep-ph, quant-ph, astro-ph, cs.CV, cs.LG, cond-mat.mes-hall, gr-qc, cond-mat.mtrl-sci, cond-mat.str-el.

hep-th stands for High Energy Physics - Theory. This category includes research papers which are centered on theoretical concepts and mathematical models in high energy physics.

hep-ph stands for High Energy Physics - Phenomenology. This category includes research papers centered on the application of theoretical physics to high energy physics experiments.

quant-ph stands for Quantum Physics. This category includes research papers centered on the theoretical and experimental aspects of the fundamental theory of quantum mechanics.

astro-ph stands for Astrophysics. This category includes research papers centered on the study of the physics of the universe, including the properties and behavior of celestial bodies.

cs.CV stands for Computer Science - Computer Vision and Pattern Recognition. This category includes research papers focused on how computers can be made to gain high-level understanding from digital images or videos.

cs.LG stands for Computer Science - Machine Learning. This category includes research papers focused on the development and implementation of algorithms that allow computers to learn from and make decisions or predictions based on data.

cond-mat.mes-hall stands for Condensed Matter - Mesoscale and Nanoscale Physics. This category includes research papers that focus on the properties and phenomena of physical systems at mesoscopic (intermediate) and nanoscopic scales.

gr-qc stands for General Relativity and Quantum Cosmology. This category includes research papers centered on theoretical and observational aspects of the theory of general relativity and its implications for understanding cosmology at the quantum scale.

cond-mat.mtrl-sci stands for Condensed Matter - Materials Science. This category includes research papers centered on the understanding, description, and development of novel materials from a physics perspective.

cond-mat.str-el stands for Condensed Matter - Strongly Correlated Electrons. This category includes research papers focused on the study of solids and liquids in which interactions among electrons play a dominant role in determining the properties of the material.

Note that you should only include the class in your reply and provide no explanations.
Please classify the following sentence into one of the 10 categories, however, if you think that the sentence could be classified into multiple categories, you may give up to 3 most likely categories:   \\ 
\bottomrule 
\end{tabular}
\label{table:prompt}
\end{table*}

In our experiment design, we assign the ground truth source of each generated text to be identical to that of the prompt sentence. 
Here, we would like to verify that our method of using the source of the prompt sentence as the ground truth source for the generated sentence is indeed a reliable approach, in addition to its benefit of simplifying the experimental evaluation.

A natural and reliable method to find the ground truth source of a generated text is to consult the opinion of human experts. Therefore, we would like to show that our method to determine the ground truth source is an accurate approximation to human evaluations. To avoid the substantial costs and resources associated with human evaluators, we have employed GPT4, noted for its human-level performance across various benchmarks~\citep{openai2023gpt4}, as a surrogate `human-like labeler'. Then, we examine whether the ground truth source determined by our method (i.e., using the source of the prompt sentence) aligns well with those determined by GPT4. 
Specifically, we use GPT4 to categorize generated texts into one of the ten ArXiv categories (i.e., data providers) using a carefully constructed prompt, as shown in Tab.~\ref{table:prompt}. After evaluating $500$ generated texts, we have found that $89.6\%$ of GPT4's decisions align with our source determination method (i.e., using the source of the prompt sentence). This validates that our method to determine the ground truth source of a generated text is a reasonable and reliable approach.

We would like to add that employing GPT4 as a `human-like labeler' is only feasible in our controlled setting here because it requires prior knowledge about all sources and detailed descriptions of the sources; see the detailed prompt in Tab.~\ref{table:prompt}. Moreover, it also incurs excessive costs in terms of monetary expenses and computations when the number of data providers is large. Therefore, we would like to clarify that this GPT4-based method is not a realistic alternative method for source attribution and is instead only employed here to verify the reliability of our method of source determination.

Additionally, note that the reason why we have used watermarked training data as the prompt sentences in our evaluation is because it leads to simple and reliable evaluations. Here, we justify this using the GPT4-based experiment as well. We use GPT4 to examine the reliability of the ground truth source determination when sentences from two held-out sets are used as the prompt sentences: when the prompt sentences are selected from unwatermarked training data and when the prompt sentences are from the validation data. The results show that when the prompt sentences are selected from unwatermarked training data, $81.6\%$ of GPT4's decisions align with the source of the prompt sentences; when the prompt sentences are from the validation data, the alignment becomes $75.0\%$. The results suggest that when the sentences from both held-out sets are used as the prompt sentences, our method to determine the ground truth source is still reasonably reliable. However, our ground truth source determination is the most reliable when sentences from watermarked training data are used as the prompt, as we have done in our main experiments. Therefore, the results justify the rationale behind our choice of using watermarked training data as prompts because it enhances the reliability of our source determination and hence the fidelity of our evaluation results.

\section{More Experimental Results}
\label{app:more:experiments:head}
\subsection{Accuracy}
\label{app:more:experiments:acc:provenance}

\subsubsection{More Details on Experimental Setup.}
In our experiments on the source attribution accuracy, 
for the ArXiv dataset, we select $50$ papers from each of the $10$ categories (App.~\ref{subsec:exp:setup:datasets}) and for every selected paper, we choose the first sentence that has been selected for watermarking (to obtain our \algllm~from second-stage pre-training of various pre-trained LLMs, see Sec.~\ref{embed_watermark} for more details on how we select the sentences for watermarking) as well as contains at least $200$ characters. Next, we use the first $200$ characters of every selected sentence (after removing the watermarks) as the input/prompt to the trained \algllm~, which generates texts with a token length of $100$. 
Similarly, for every book (i.e., data provider) in the BookSum dataset, we select the first $50$ sentences that have been selected for watermarking as well as have at least $200$ characters.
As a result, for both datasets, we have selected $50$ sentences to be used as the inputs/prompts to our \algllm, which corresponds to $50$ trials of source attribution for each of the $10$ data providers. In addition, the source attribution accuracy and F1 score for OPT-1.3B model are presented in App.~\ref{app:more:exp:scalability}, together with the scalability results.

\subsubsection{F1 score.} In our main experiments, we have reported the macro F1 score for a more comprehensive evaluation. To compute the F1 score, here we first define precision as the number of correct watermarks (watermarks that correctly correspond to its true source) for the data provider $i$ divided by the number of all generated watermarks that correspond to the data provider $i$ and define recall as the number of correct watermarks divided by the number of trails of the data provider $i$. We calculate the precision and recall for each data provider and obtain $precision_{i}$ and $recall_{i}$. Subsequently, We obtain $precision_{ma}$ and $recall_{ma}$ by averaging the precisions and recalls from all data providers. 
Therefore, the macro F1 score can be computed as:

\begin{equation}
F_1 = 2 \times \frac{precision_{ma} \times recall_{ma}}{precision_{ma} + recall_{ma}}.
\label{eq:f1}
\end{equation}

\subsubsection{BM25 as source attribution baseline.}
\label{app:more:experiments:acc:BM25}
BM25 is a well-known search engine algorithm that can potentially be utilized to perform source attribution given the generated sentences. 
In our experiments, we have implemented the BM25 from GitHub \footnote{\url{https://github.com/dorianbrown/rank_bm25}} as a source attribution baseline for comparison. Specifically, we apply BM25 and take the unwatermarked training data as the corpus, and take the same generated sentences from our \algllm~(the watermarks are cleaned) as input. Subsequently, we can use BM25 to find the top-$k$ closest data providers in the training data. BM25 operates as a post-hoc process, which may slow down source identification, especially for larger number of potential sources.

\subsubsection{Source Attribution Accuracy for Each Data Provider.}
\label{app:more:experiments:acc:providers}
Tabs.~\ref{table:accuracy:providers:arxiv} and~\ref{table:accuracy:providers:booksum} show the detailed results on source attribution accuracy and F1 score for the $10$ different data providers, in addition to Tab.~\ref{table:accuracy:gpt} in Sec.~\ref{subsec:exp:accuracy}. The results show that the accuracy remains balanced across the data providers. 

\begin{table*}
\caption{
Source attribution accuracy and F1 score achieved by our \algllm~(i.e., obtained from second-stage pre-training of different models on various datasets) for the ArXiv dataset.
}
\centering
\resizebox{\linewidth}{!}{
\begin{tabular}{l|ccc|ccc|ccc}
\toprule
\multirow{2}{*}{Data Provider} & \multicolumn{3}{c|}{GPT2}& \multicolumn{3}{c|}{OPT}&\multicolumn{3}{c}{Llama2}\\
& acc. & top-$3$. & F1 & acc. & top-$3$. & F1 & acc. & top-$3$. & F1 \\ 
\midrule
hep-th   &    $65.60_{\pm7.40}$ &  $94.40_{\pm2.61}$   &  $0.730_{\pm0.04}$   &    $67.60_{\pm13.22}$ &  $99.20_{\pm1.10}$   &  $0.622_{\pm0.35}$     &     $88.00_{\pm5.29}$ &  $96.67_{\pm3.06}$   &  $0.810_{\pm0.07}$  \\ 
hep-ph    &    $85.20_{\pm4.15}$  &  $96.80_{\pm3.03}$   &  $0.708_{\pm0.13}$  &    $87.60_{\pm5.55}$  &  $98.80_{\pm2.68}$   &  $0.820_{\pm0.07}$   &     $71.33_{\pm8.08}$ &  $96.67_{\pm2.31}$   &  $0.853_{\pm0.08}$ \\ 
quant-ph  &   $74.80_{\pm6.72}$  &  $91.60_{\pm5.90}$    &  $0.678_{\pm0.08}$  &   $76.80_{\pm6.72}$  &  $98.00_{\pm3.46}$     &  $0.808_{\pm0.07}$    &     $72.00_{\pm5.29}$ &  $95.33_{\pm1.15}$     &  $0.820_{\pm0.13}$   \\ 
astro-ph  &   $86.40_{\pm2.61}$   & $94.40_{\pm2.61}$    &  $0.793_{\pm0.03}$ &   $86.00_{\pm4.47}$    & $98.40_{\pm2.19}$    &  $0.818_{\pm0.03}$   &     $69.33_{\pm6.43}$ &  $98.00_{\pm2.00}$   &  $0.850_{\pm0.06}$ \\ 
cs.CV   &   $82.00_{\pm4.00}$  & $95.20_{\pm3.03}$   &  $0.790_{\pm0.08}$  &   $85.20_{\pm6.72}$  & $99.20_{\pm1.10}$     &  $0.610_{\pm0.35}$  &     $78.00_{\pm2.00}$ &  $97.33_{\pm2.31}$   &  $0.787_{\pm0.10}$  \\ 
cs.LG    &  $77.60_{\pm3.58}$  &  $98.80_{\pm1.10}$    &  $0.808_{\pm0.08}$    &  $83.20_{\pm4.38}$   &  $99.60_{\pm0.89}$   &  $0.688_{\pm0.06}$  &     $79.33_{\pm1.15}$ &  $98.00_{\pm2.00}$   &  $0.737_{\pm0.06}$  \\ 
cond-mat.mes-hall  & $64.80_{\pm5.22}$ &  $98.40_{\pm0.89}$   &  $0.693_{\pm0.08}$  &  $74.00_{\pm3.74}$ & $99.20_{\pm1.10}$   &  $0.742_{\pm0.10}$  &     $76.00_{\pm8.72}$ &  $99.33_{\pm1.15}$  &  $0.783_{\pm0.10}$  \\ 
gr-qc   &   $76.40_{\pm2.61}$  & $96.40_{\pm1.67}$   &  $0.748_{\pm0.08}$  &   $82.00_{\pm5.10}$ & $99.20_{\pm1.10}$    &  $0.728_{\pm0.09}$  &     $86.00_{\pm5.29}$ &  $98.00_{\pm2.00}$    &  $0.780_{\pm0.14}$ \\ 
cond-mat.mtrl-sci &    $64.80_{\pm3.63}$   & $95.20_{\pm3.35}$  &  $0.845_{\pm0.06}$  &   $71.60_{\pm5.18}$  & $99.20_{\pm1.79}$   &  $0.752_{\pm0.11}$  &     $73.33_{\pm6.43}$ &  $94.00_{\pm5.29}$    &  $0.860_{\pm0.06}$ \\ 
cond-mat.str-el    &  $70.80_{\pm1.01}$  & $96.40_{\pm1.67}$    &  $0.810_{\pm0.11}$ &  $69.60_{\pm8.29}$ &   $99.60_{\pm0.89}$   &  $0.752_{\pm0.11}$  &     $80.67_{\pm2.31}$ &  $96.00_{\pm4.00}$   &  $0.703_{\pm0.04}$  \\ 
\midrule
Overall   & $74.84_{\pm10.06}$  & $95.76_{\pm1.67}$   &  $0.758_{\pm0.02}$  & $78.36_{\pm8.29}$ & $99.04_{\pm0.89}$   &  $0.738_{\pm0.05}$  &     $77.40_{\pm1.91}$ &  $96.87_{\pm1.62}$    &  $0.800_{\pm0.03}$  \\ 
\bottomrule 
\end{tabular}
}
\label{table:accuracy:providers:arxiv}
\end{table*}

\begin{table*}
\caption{
Source attribution accuracy and F1 score achieved by our \algllm~(i.e., obtained from second-stage pre-training of different models on various datasets) for BookSum dataset.
}
\centering
\resizebox{\linewidth}{!}{
\begin{tabular}{l|ccc|ccc|ccc}
\toprule
\multirow{2}{*}{Data Provider} & \multicolumn{3}{c|}{GPT2}& \multicolumn{3}{c|}{OPT}&\multicolumn{3}{c}{Llama2}\\
& acc. & top-$3$. & F1  & acc. & top-$3$. & F1 & acc. & top-$3$. & F1 \\ 
\midrule
Adam Bede   &    $82.40_{\pm3.29}$ &  $95.60_{\pm2.19}$    &  $0.805_{\pm0.01}$  &     $85.20_{\pm3.35}$ &  $96.00_{\pm2.15}$   &  $0.745_{\pm0.01}$  &  $85.33_{\pm5.03}$  &  $94.67_{\pm6.11}$    &  $0.820_{\pm0.06}$ \\ 
David Copperfield    &  $80.00_{\pm6.63}$  &  $88.40_{\pm5.90}$   &  $0.670_{\pm0.04}$  &    $77.20_{\pm6.72}$  &  $91.60_{\pm1.67}$   &  $0.820_{\pm0.03}$  &  $80.67_{\pm2.31}$  &  $96.67_{\pm2.31}$   &  $0.755_{\pm0.28}$  \\ 
Dracula  &   $66.80_{\pm6.26}$  &  $86.00_{\pm6.16}$    &  $0.880_{\pm0.10}$  &   $71.60_{\pm8.17}$  &  $91.60_{\pm2.97}$   &  $0.905_{\pm0.12}$   &  $74.67_{\pm6.11}$  &  $90.67_{\pm4.16}$   &  $0.915_{\pm0.06}$   \\ 
Hamlet  &  $91.20_{\pm4.38}$   & $96.80_{\pm2.28}$   &  $0.700_{\pm0.08}$  &   $97.60_{\pm2.19}$   & $99.20_{\pm1.10}$   &  $0.920_{\pm0.10}$  &  $98.00_{\pm0.00}$  &  $99.33_{\pm1.15}$   &  $0.810_{\pm0.03}$  \\ 
Henry IV Part 1   &   $90.40_{\pm2.61}$  & $98.40_{\pm2.61}$   &  $0.375_{\pm0.53}$ &   $97.20_{\pm1.10}$   & $99.60_{\pm0.89}$   &  $0.885_{\pm0.13}$   &  $98.67_{\pm1.15}$  &  $100.00_{\pm0.00}$   &  $0.995_{\pm0.01}$ \\ 
Ivanhoe    & $83.60_{\pm3.28}$ &  $94.40_{\pm1.67}$   &  $0.790_{\pm0.21}$  &  $89.20_{\pm5.40}$   &  $93.60_{\pm4.34}$   &  $0.920_{\pm0.04}$    &  $85.33_{\pm8.33}$  &  $94.67_{\pm4.16}$    &  $0.820_{\pm0.08}$  \\ 
Jane Eyre  & $74.00_{\pm6.16}$ &  $90.00_{\pm4.00}$ &  $0.805_{\pm0.11}$ &  $80.00_{\pm2.00}$     & $96.40_{\pm3.85}$  &  $0.810_{\pm0.10}$  &  $77.33_{\pm15.53}$    &  $94.67_{\pm3.06}$     &  $0.785_{\pm0.18}$      \\ 
Little Women   &   $85.60_{\pm2.61}$  & $94.00_{\pm3.16}$   &  $0.650_{\pm0.10}$  &   $94.00_{\pm3.16}$ & $98.00_{\pm2.00}$   &  $0.820_{\pm0.07}$   &  $92.67_{\pm5.77}$  &  $100.00_{\pm0.00}$   &  $0.815_{\pm0.02}$ \\ 
Middlemarch &    $72.80_{\pm3.35}$   & $94.40_{\pm2.61}$  &  $0.755_{\pm0.09}$  &  $76.00_{\pm5.83}$ & $93.20_{\pm3.35}$   &  $0.755_{\pm0.06}$  &  $74.67_{\pm7.02}$  &  $93.33_{\pm4.62}$   &  $0.815_{\pm0.02}$  \\ 
The Pickwick Papers    &  $52.40_{\pm4.78}$  &  $80.00_{\pm6.16}$   &  $0.775_{\pm0.11}$  &  $64.00_{\pm9.27}$ &   $79.20_{\pm5.76}$   &  $0.850_{\pm0.21}$ &  $65.33_{\pm6.43}$  &  $88.67_{\pm1.15}$   &  $0.850_{\pm0.21}$  \\ 
\midrule
Overall   & $77.92_{\pm1.57}$  & $91.80_{\pm0.24}$   &  $0.723_{\pm0.08}$  & $83.20_{\pm1.08}$ & $93.84_{\pm1.01}$   &  $0.840_{\pm0.01}$ &  $83.27_{\pm4.50}$  &  $95.27_{\pm1.53}$   &  $0.840_{\pm0.06}$  \\ 
\bottomrule 
\end{tabular}
}
\label{table:accuracy:providers:booksum}
\end{table*}



\subsubsection{Fine-grained Error Analysis of Source Attribution.}
Tab.~\ref{table:error:analysis} shows the errors of misclassification and incorrect watermark, as mentioned in Sec.~\ref{subsec:exp:accuracy}. The results show that most source attribution errors are caused by generated texts 
exhibiting the characteristics of multiple data providers. 


\begin{table*}[t]
\caption{Error analysis of watermarks incurred by our \algllm~that is obtained from second-stage pre-training of the GPT2 model on the ArXiv dataset. Note that the numbers shown here are the average taken across $5$ runs with different random seeds and `wtm' is the short form of ``watermark''. }
\centering
\resizebox{0.55\linewidth}{!}{
\begin{tabular}{l|cccc}
\toprule
category    & n\_wtm & n\_match & misclassify & incorrect  \\ 
\midrule
hep-th     & $50$  & $32.8_{\pm3.72}$   & $17.2_{\pm3.72}$ & $0_{\pm0.00}$      \\ 
hep-ph     & $50$  & $42.6_{\pm2.07}$   &  $7.4_{\pm2.07}$ & $0_{\pm0.00}$   \\ 
quant-ph   & $50$  & $37.4_{\pm3.36}$  & $12.6_{\pm3.36}$ & $0_{\pm0.00}$            \\ 
astro-ph   & $50$  &  $43.2_{\pm1.30}$  &   $6.8_{\pm1.30}$ & $0_{\pm0.00}$    \\ 
cs.CV      & $50$  & $41.0_{\pm2.00}$ & $9.0_{\pm2.00}$ & $0_{\pm0.00}$ \\ 
cs.LG      & $50$  &  $38.8_{\pm1.79}$  & $11.2_{\pm1.79}$ & $0_{\pm0.00}$         \\ 
cond-mat.mes-hall  & $50$ &  $32.4_{\pm2.61}$ &  $17.6_{\pm2.61}$ & $0_{\pm0.00}$         \\ 
gr-qc  & $50$  & $38.2_{\pm1.30}$ &   $11.8_{\pm1.30}$ & $0_{\pm0.00}$    \\ 
cond-mat.mtrl-sci   & $50$  & $32.4_{\pm1.82}$ & $17.6_{\pm1.82}$  & $0_{\pm0.00}$   \\ 
cond-mat.str-el  & $50$  &  $35.4_{\pm5.03}$ & $14.6_{\pm5.03}$ & $0_{\pm0.00}$ \\ 
\midrule
Total      & $500$ & $374.2_{\pm10.18}$ & $125.8_{\pm10.18}$ & $0_{\pm0.00}$  \\ 
\bottomrule
\end{tabular}
}
\label{table:error:analysis}
\end{table*}

\subsubsection{Data Provenance.}
\label{app:data:provenance}
We show here that \alg's ability to perform reliable source attribution also allows us to achieve accurate data provenance.
Since the data providers are given both their own unique watermarks (Sec.~\ref{embed_watermark}) and the watermark decoder, they can request their \emph{data provenance}. 
Specifically, when a data provider requests data provenance, it uses its own text data (without watermark) as the input/prompt to our trained \algllm~to verify whether the generated watermark 
matches its own (Fig.~\ref{fig: workflow}).
We consider $20$ categories/data providers in the ArXiv dataset, including $10$ categories whose data was used for second-stage pre-training of GPT2 to obtain \algllm~and $10$ other categories whose data was not used.
We select $50$ papers from each category 
and choose a sentence from every selected paper to use as the input/prompt to \algllm~for generating a watermark.
The results in Tab.~\ref{table: provenance} show that for the first $10$ categories whose data was \emph{not used} to obtain \algllm, we are consistently able to recognize that their data was not misused; for the other $10$ categories whose data \emph{was used} to obtain \algllm, we can also identify this
with high accuracy of $74.84\%$ and top-$3$ accuracy of $95.76\%$. The results show that, due to its ability to perform accurate source attribution, our \alg~framework can also achieve reliable data provenance.

\subsubsection{More Diverse Datasets}
\label{app:diverse:datasets}
To verify the generalizability of our \alg~framework on more diverse datasets from various domains, including those that are potentially less curated and less formal, we have adopted several additional datasets from other domains and selected $10$ data providers for our experiment, including Wikipedia, news, and movie reviews. To elaborate, the additional datasets we consider are:

\noindent\textbf{DBpedia14}~\citep{NIPS2015_250cf8b5} is an ontology classification dataset taken from DBpedia $2014$, containing $14$ classes and $560$k training samples. The content is extracted from information created in Wikipedia. In our experiments, we refer to the `title' column, which denotes the ontology class of the content, to categorize the data providers. 

\noindent\textbf{CC-News}~\citep{Hamborg2017} is a representative less-curated and less-formal dataset. 
It contains approximately $700$K English language news articles sourced from various global news sites. The dataset is collected by crawling the news websites for main text content. 
Importantly, \emph{no additional preprocessing is conducted} on the text content, resulting in a dataset that is less curated, quite noisy, and may include diverse elements such as different languages, emojis, URLs, Unicode, etc. In our experiments, we categorize data providers based on the `domain' column, which denotes the distinct news media.

\noindent\textbf{IMDB62}~\citep{seroussi2014authorship} comprises movie reviews written by $62$ distinct authors, with each author serving as an individual data provider. Each author contributes $1,000$ reviews, which are sampled from their complete collection of reviews. This dataset facilitates the evaluation of our approach in a context where the texts share similar thematic content. The dataset is relatively noisy, as it may include spelling and grammatical errors. In our experiments, we categorize data providers based on the `userId' column. Note that specifically for this dataset, since each data provider contributes too few data samples, we perform $10$ epochs of second-stage pretraining to obtain our \algllm~.

\noindent\textbf{Fake News Opensources}\footnote{\url{https://huggingface.co/datasets/andyP/fake_news_en_opensources}} comprises $8,529,090$ individual articles, which were scraped from various news websites between late 2017 and early 2018, encompassing a total of $647$ distinct sources. Similar to the CC-News dataset, this dataset is less curated. We categorize the data providers based on the 'domain' column, which specifies the distinct news media sources.



The source attribution accuracy on these more diverse datasets using our \algllm~adopting Llama2 as the pre-trained model is illustrated in Tab.~\ref{table: diverse dataset}. The results indicate that our framework consistently achieves decent accuracy in source attribution across various datasets that mostly remain higher than the BM25 baseline. This further verifies the effectiveness of our \alg~framework on various datasets. 
However, it is also observed that the accuracy tends to be lower on the less curated and noisy datasets (i.e., CC-News, IMDB62, and Fake News) compared to the datasets with more formal language (i.e., ArXiv, BookSum, DBpedia14). 


\begin{table}[ht]
\caption{
Source attribution accuracy on the dataset from diverse domains. 
}
\centering
\begin{tabular}{c|cccccc}
\toprule
 \multirow{2}{*}{Dataset} &  \multicolumn{2}{c}{acc.}   & \multicolumn{2}{c}{top-$3$.} & \multicolumn{2}{c}{top-$5$.} \\
 &  BM25 & WASA & BM25 & WASA & BM25 & WASA \\
\midrule
DBpedia14 & $86.00$ & $\textbf{90.80}$ & $\textbf{96.00}$ & $93.20$ & $\textbf{98.20}$ &  $94.00$  \\ 
CC-News & $45.00$ & $\textbf{60.20}$ & $71.20$ & $\textbf{79.40}$  & $84.00$ &  $\textbf{85.00}$  \\ 
IMDB62 & $29.60$ & $\textbf{67.20}$ & $48.20$ & $\textbf{89.40}$  & $65.80$ &  $\textbf{97.00}$  \\ 
FakeNews & $33.40$ & $\textbf{62.63}$ & $53.40$ & $\textbf{85.00}$  & $62.20$ &  $\textbf{93.13}$  \\ 
\bottomrule 
\end{tabular}
\label{table: diverse dataset}
\end{table}

\subsubsection{Analysis of Data Sources}
In Sec.~\ref{sec:intro}, we have mentioned that we consider data providers that contribute balanced data with unique characteristics. Here we analyze and show the balance and unique characteristics of the data sources in each dataset we have adopted in Tab.~\ref{table:balance}. Firstly, we calculate the imbalance ratio by dividing the number of tokens in the largest data source by that in the smallest, hence larger imbalance ratio suggests that the data sources are more imbalanced. The results shown in Table~\ref{table:balance} indicate that the data sources in our adopted datasets are not perfectly balanced while some are particularly imbalanced. This indicates that our proposed method can generalize to imbalanced data sources and achieve decent source attribution accuracy.

Our datasets also encompass a variety of unique characteristics, which ensures that our framework is applicable across different applications. These include academic fields (ArXiv), general knowledge (DBpedia14), and attributing authorship based on story or writing style (BookSum, CC-News, IMDB62, FakeNews). Our analysis reveals that both our framework and baselines face challenges in scenarios where the distinguishing features are restricted to writing style and word choice, naturally resulting in lower accuracy. This underscores the inherent difficulties of source attribution in homogeneous topic environments, yet our method consistently outperforms the baselines across these challenging conditions.

\begin{table}[h]
\caption{Balance and unique characteristics of the data sources in each dataset. 
}
\centering
\begin{tabular}{l|cc}
\toprule
Dataset & balance & Characteristics \\
\midrule
ArXiv & 2.5 & academic knowledge fields (with overlaps)\\
BookSum & 17.51 & book stories and writing style from book authors \\
DBpedia14 & 1.66 & common knowledge fields \\
CC-News & 5.37 & writing style and word choices from news publishers \\
IMDB62 & 1.64 & writing style and word choices from common Internet users \\
FakeNews & 25.45 & writing style and word choices from news publishers \\
\bottomrule 
\end{tabular}
\label{table:balance}
\end{table}

\begin{table}[h]
\caption{Reliable data provenance can be achieved due to the ability of \algllm~to perform accurate source attribution. \algllm~is obtained from second-stage pre-training of the GPT2 model on the ArXiv dataset. Note that the numbers shown here are the average taken across $5$ runs with different random seeds. `wtm' is the short form of ``watermark''. 
}
\centering
\begin{tabular}{l|cc}
\toprule
category    &  n\_wtm & data provenance (n\_match)  \\ 
\midrule
cond-mat.soft    & $50$     &    \ding{55}  ($0_{\pm0.00}$)  \\ 
q-bio.PE  & $50$    &   \ding{55}  ($0_{\pm0.00}$)     \\ 
cs.SY   & $50$    &     \ding{55}   ($0_{\pm0.00}$)                \\ 
eess.IV   & $50$     &    \ding{55}   ($0_{\pm0.00}$)      \\ 
hep-ex     & $50$        &     \ding{55}   ($0_{\pm0.00}$)      \\ 
math.LO    & $50$      &      \ding{55}  ($0_{\pm0.00}$)       \\ 
math.NA & $50$    &      \ding{55}   ($0_{\pm0.00}$)      \\ 
math.ST   & $50$       &    \ding{55} ($0_{\pm0.00}$)       \\ 
nlin.SI & $50$      &   \ding{55}  ($0_{\pm0.00}$)            \\ 
physics.class-ph & $50$     &    \ding{55}   ($0_{\pm0.00}$)  \\
hep-th  & $50$     &      \ding{51}  ($32.8_{\pm3.70}$)      \\ 
hep-ph     & $50$     &     \ding{51} ($42.6_{\pm2.07}$)            \\ 
quant-ph   & $50$     &      \ding{51} ($37.4_{\pm3.36}$)        \\ 
astro-ph   & $50$       &     \ding{51} ($43.2_{\pm1.30}$)        \\ 
cs.CV      & $50$        &   \ding{51} ($41.0_{\pm2.00}$)     \\ 
cs.LG      & $50$       &   \ding{51} ($38.8_{\pm1.79}$)    \\ 
cond-mat.mes-hall  & $50$      &   \ding{51} ($32.4_{\pm2.61}$)    \\ 
gr-qc      & $50$        &   \ding{51} ($38.2_{\pm1.30}$)      \\ 
cond-mat.mtrl-sci   & $50$      &    \ding{51} ($32.4_{\pm1.82}$)    \\ 
cond-mat.str-el  & $50$    &  \ding{51} ($35.4_{\pm5.03}$)     \\
\bottomrule 
\end{tabular}
\label{table: provenance}
\end{table}

\subsection{Robustness}
\label{app:more:exp:robustness}

\subsubsection{Additional Attacks on Generated Sentences with Embedded Watermarks}
\label{app:robustness:type:1}

As discussed in Sec.~\ref{robustness}, an adversary may \emph{additionally modify the content of the generated sentence} while removing/modifying the generated watermarks. Here, we will consider insertion, deletion, synonym substitution, and syntactic transformation attacks.
In \textbf{insertion attacks} on a generated watermarked sentence, either one word is randomly inserted into the sentence (i.e., \emph{localized insertion attacks}), or various words are randomly interspersed throughout the sentence (i.e., \emph{dispersed insertion attacks})~\citep{kamaruddin2018review}.
For dispersed insertion attacks, we vary the attack strengths by changing the number of inserted words from $5\%$ to $20\%$ of the total number of words in the sentence.
In \textbf{deletion attacks},
some words in the text are randomly deleted.
In \textbf{synonym substitution attacks}~\citep{kamaruddin2018review}, an adversary substitutes some words in the generated sentence with their synonyms while preserving the semantic meaning of the sentence.
We again test different attack strengths by varying the percentage of randomly deleted and substituted words.
In addition, we also performed the \textbf{syntactic transformation attack} on the generated sentences whereby an adversary transforms the sentences (without altering their semantic meanings) via techniques such as modifying the prepositions, tenses, 
and other syntax components.
Here, we adopt a strong variant of such attacks,  
which paraphrases the input sentence using 
the PEGASUS model fine-tuned for paraphrasing \citep{zhang2020pegasus}. The accuracy (top-$3$ accuracy) after the syntactic transformation attacks is $66.28\%$ ($89.56\%$). 
Besides the above attacks, we have further considered a more recent oracle-based attack as proposed in~\citep{zhang2023watermarks}, which generates perturbation oracles with an open-source model and removes the watermarks in the attacked sentence. Under this attack, the watermark regeneration defense is also performed and we are still able to achieve a source attribution accuracy of $75.80\%$, which further validates the robustness of our \alg~framework.
The \textbf{robustness} of our \alg~framework can be validated by the marginal performance degradation in Tab.~\ref{table: attacks}. In addition, the standard deviations for this part of the results in Tab.~\ref{table: attacks} are reported in Tab.~\ref{table: attacks_std}.

\subsubsection{Attacks on Input Sentences (Prompts)}
\label{app:more:exp:robustness:input}
\label{app:robustness:type:2}
An adversary may also 
manipulate the input sentence (prompt) to our trained \algllm~to disrupt watermark generation and hence source attribution. 
The \textbf{insertion, deletion, and syntactic transformation attacks} are the same as those described in App.~\ref{app:robustness:type:1}, except that these attacks are performed on the input sentences here.
Similar to App.~\ref{app:robustness:type:1}, we vary the attack strengths for these three types of attacks.
The results in 
Tab.~\ref{table: attacks} show that these attacks also only lead to marginal degradation in the source attribution accuracy.
Moreover, under the strong syntactic transformation attacks, the source attribution remains accurate (with an accuracy of $63.00\%$ and a top-$3$ accuracy of $89.00\%$), which provides further evidence for the robustness of our \alg~framework against attacks on the input sentences.
Its robustness against these attacks can again be explained by its reliable texts-to-watermarks mapping,
which allows our \algllm~to consistently generate the correct watermarks even if the prompt is perturbed. The standard deviations for this part of the results in Tab.~\ref{table: attacks} are reported in Tab.~\ref{table: attacks_std_input}.

\begin{table*}[t]
\caption{
Source attribution accuracy and standard deviation using regenerated watermarks by \algllm~(from second-stage pre-training of GPT2 on ArXiv dataset) under attacks on \textbf{generated sentences with embedded watermarks}  (\emph{in addition to watermark removal/modification attacks}).
}
\centering
\resizebox{0.73\linewidth}{!}{
\begin{tabular}{l|cc|cc|cc}
\toprule
\multirow{3}{*}{strength}  & \multicolumn{6}{c}{attacks on generated sentences with embedded watermarks}  \\

&\multicolumn{2}{c|}{insertion attack} & \multicolumn{2}{c|}{deletion attack} & \multicolumn{2}{c}{synonym substitution} \\

& acc.  & top-$3$.  & acc. & top-$3$. & acc. & top-$3$. \\
\midrule 
$0\%$  & $71.60_{\pm1.33}$  & $93.76_{\pm0.57}$ & $71.60_{\pm1.33}$ & $93.76_{\pm0.57}$ & $71.60_{\pm1.33}$ & $93.76_{\pm0.57}$ \\
Localized    & $71.40_{\pm0.89}$  & $93.56_{\pm0.46}$  & -  &  -  &-&- \\
$5\%$  & $70.12_{\pm1.35}$ & $93.20_{\pm0.14}$  & $71.08_{\pm0.92}$  & $93.92_{\pm0.66}$ &  $70.52_{\pm0.83}$ & $93.52_{\pm0.64}$  \\
$10\%$  &  $69.12_{\pm1.90}$ &  $92.20_{\pm0.47}$  & $71.84_{\pm1.36}$  & $93.68_{\pm0.78}$ &  $71.02_{\pm0.81}$ & $92.88_{\pm0.95}$  \\
$15\%$  &  $66.92_{\pm1.32}$  &  $91.96_{\pm0.91}$ & $71.36_{\pm1.01}$  & $94.04_{\pm0.79}$ & $70.96_{\pm0.52}$  & $92.72_{\pm0.46}$  \\
$20\%$  &  $65.12_{\pm2.37}$  &  $91.44_{\pm0.50}$  & $70.00_{\pm1.17}$ & $93.24_{\pm0.54}$ & $69.20_{\pm1.89}$ & $93.20_{\pm0.62}$  \\
\bottomrule
\end{tabular}
}
\label{table: attacks_std}
\end{table*}

\begin{table*}[t]
\caption{
Source attribution accuracy and standard deviation using regenerated watermarks by \algllm~(from second-stage pre-training of GPT2 on ArXiv dataset) under attacks on \textbf{input sentences}  (\emph{in addition to watermark removal/modification attacks}).
}
\centering
\resizebox{0.7\linewidth}{!}{
\begin{tabular}{l|cc|cc|cc}
\toprule
\multirow{3}{*}{strength} & \multicolumn{6}{c}{attacks on input sentences} \\

&\multicolumn{2}{c|}{insertion attack} & \multicolumn{2}{c|}{deletion attack} & \multicolumn{2}{c}{synonym substitution}\\

& acc.  & top-$3$.  & acc. & top-$3$. & acc. & top-$3$.  \\
\midrule 
$0\%$  & $74.84_{\pm2.04}$ & $95.76_{\pm1.24}$ & $74.84_{\pm2.04}$ & $95.76_{\pm1.24}$ & $74.84_{\pm2.04}$ & $95.76_{\pm1.24}$ \\
Localized  & $74.20_{\pm1.76}$ &  $95.40_{\pm1.02}$ & -  & - & - & - \\
$5\%$  & $74.20_{\pm2.40}$ &  $95.40_{\pm0.62}$ & $73.56_{\pm1.48}$ & $95.52_{\pm0.86}$ & $72.84_{\pm2.13}$ & $95.24_{\pm1.06}$ \\
$10\%$   &  $72.88_{\pm2.74}$ &   $94.68_{\pm1.17}$ & $72.96_{\pm2.05}$ &  $94.68_{\pm0.87}$  & $73.60_{\pm1.84}$ & $95.00_{\pm1.09}$  \\
$15\%$   &  $71.52_{\pm2.09}$ &  $93.20_{\pm0.71}$ & $72.68_{\pm1.74}$ & $94.12_{\pm1.02}$ & $71.88_{\pm1.40}$ & $94.20_{\pm1.10}$ \\
$20\%$  &   $68.60_{\pm1.36}$   &  $93.40_{\pm0.55}$ & $72.68_{\pm2.73}$ &  $94.12_{\pm1.45}$   &  $72.08_{\pm1.09}$ & $93.76_{\pm0.52}$ \\
\bottomrule
\end{tabular}
}
\label{table: attacks_std_input}
\end{table*}

\begin{table*}[ht]
\caption{
Source attribution accuracy using regenerated watermarks by \algllm~(from second-stage pre-training of GPT2 on ArXiv dataset) under character-level attacks on generated sentences with embedded watermarks  (\emph{in addition to watermark removal/modification attacks}).
}
\centering
\resizebox{0.8\linewidth}{!}{
\begin{tabular}{l|cc|cc||c|cc}
\toprule
\multirow{2}{*}{strength}  & 
\multicolumn{2}{c|}{insertion attack} & \multicolumn{2}{c||}{deletion attack} & \multirow{2}{*}{strength}  & \multicolumn{2}{c}{swap attack}\\
& acc.  & top-$3$.  & acc. & top-$3$. && acc. & top-$3$. \\
\midrule 
$0\%$  & $71.60_{\pm1.33}$  & $93.76_{\pm0.57}$   & $71.60_{\pm1.33}$  & $93.76_{\pm0.57}$ & $0\%$ & $71.60_{\pm1.33}$  & $93.76_{\pm0.57}$ \\
$5\%$  & $69.60_{\pm2.05}$   & $91.08_{\pm1.79}$   & $69.60_{\pm2.03}$   & $92.08_{\pm1.85}$ & $2\%$ & $69.90_{\pm6.48}$ & $91.88_{\pm2.65}$ \\
$10\%$  &  $60.95_{\pm3.21}$ &  $89.64_{\pm4.73}$   & $60.15_{\pm2.75}$  & $88.96_{\pm5.08}$ & $4\%$ & $68.70_{\pm8.77}$  & $91.28_{\pm3.11}$  \\
\bottomrule
\end{tabular}
}
\label{table: character attacks}
\end{table*}

\subsubsection{Character-Level Attacks}
Apart from the word-level attacks that \emph{additionally modify the content of the generated sentence} while removing/modifying the generated watermarks, 
for the regenerated watermarks,
we would also like to explore some character-level attacks on the generated sentences similar to the setting in the work of~\citet{8424632}. These attacks aim to disrupt the original texts at a character level, thus making them stronger than word-level attacks; however, it is also potentially easier to identify such attacks~\citep{LI2023119170}. Specifically, we consider character-level insertion, deletion, and character-swapping attacks. We also adopt our regeneration defense after these attacks are applied. Tab.~\ref{table: character attacks} shows the source attribution accuracy for the regenerated watermarks.

As shown in Tab.~\ref{table: character attacks}, under these strong character-level attacks, the source attribution accuracy of our watermarks is lowered yet remains decent. In addition, we would like to clarify that since these character-level attacks can heavily influence the original readability of the texts, their feasibility in realistic scenarios may be limited.

\subsection{Scalability}
\label{app:more:exp:scalability}
In Sec.~\ref{scalability}, we have verified \alg's scalability to a large number of data providers using the ArXiv dataset. Here, we will also show in Tab.~\ref{table: opt} the attribution accuracy obtained from the OPT model and in Tab.~\ref{table: scalibility_booksum} the source attribution accuracy for a larger number of books (i.e., data providers) using the BookSum dataset. 
It can be observed that \alg~generally does not scale as well (especially for GPT2 and OPT) on the BookSum dataset as compared to the ArXiv dataset because each data provider in the former offers much less data. 
It is also noteworthy that the larger Llama2 model produces higher accuracy than the smaller GPT2 and OPT models, especially when the number of providers is larger on the BookSum dataset. 
Nevertheless, the source attribution accuracy still remains relatively high compared with BM25.
As mentioned in Sec.~\ref{scalability}, with more data providers, we recommend using $k>1$ in top-$k$ source attribution due to higher resulting accuracy and identifying the true source from among them.

For an even larger number of data providers, we adopt the 
\noindent\textbf{Reddit Webis-TLDR-17}~\citep{volske-etal-2017-tl} dataset, which comprises $3,848,330$ posts, each with an average length of $270$ words. These posts originate from various subreddits created by different users. Although the dataset was initially developed for summarization tasks, we utilize only the 'body' column for the text and the 'subreddit' column to identify the data providers. Using this dataset, we consider $500$ data providers. Table~\ref{table: reddit acc} shows the source attribution accuracy when the number of data providers increases to $500$, where the accuracy still remains high compared with the BM25 baseline.

\begin{table}[h]
\caption{
Source attribution accuracy and F1 score for OPT-1.3B model on ArXiv dataset. 
}
\centering
\begin{tabular}{c|cccc}
\toprule
n & acc.   & top-$3$. & top-$5$. & F1 \\
\midrule
$10$   & $78.36 \pm_{2.04}$ & $99.04 \pm_{1.22}$ & $99.36 \pm_{0.61}$  & $0.743 \pm_{0.06}$ \\ 
$25$  & $69.76 \pm_{0.21}$ & $90.48 \pm_{0.71}$ & $95.76 \pm_{0.79}$  & $0.697 \pm_{0.01}$ \\ 
$50$  & $61.14 \pm_{1.37}$ & $82.63 \pm_{1.25}$ & $89.37 \pm_{0.82}$  & $0.613 \pm_{0.01}$ \\ 
$100$ & $48.86 \pm_{0.95}$ & $73.34 \pm_{0.76}$  & $81.54 \pm_{0.27}$  & $0.487 \pm_{0.01}$ \\
\bottomrule 
\end{tabular}
\label{table: opt}
\end{table}

\begin{table*}[h]
\caption{
Source attribution accuracy for different numbers of books (i.e., data providers) on the  BookSum dataset.
}
\centering
\resizebox{1.00\linewidth}{!}{
\begin{tabular}{cc|ccc|ccc|ccc}
\toprule
\multirow{2}{*}{n} & \multirow{2}{*}{BM25} & \multicolumn{3}{c|}{GPT2}& \multicolumn{3}{c|}{OPT}& \multicolumn{3}{c}{Llama2}\\
 &  & acc.   & top-$3$. & top-$5$. & acc.   & top-$3$. & top-$5$. & acc.   & top-$3$. & top-$5$. \\
\midrule
$10$  & $54.07_{\pm12.3}$  & $77.92_{\pm1.57}$ & $91.80_{\pm0.24}$ &  $96.52_{\pm0.76}$   & $83.20_{\pm1.08}$ & $93.84_{\pm1.01}$ & $97.80_{\pm0.42}$ & $83.27_{\pm4.50}$ & $95.27_{\pm1.53}$ & $97.67_{\pm0.46}$ \\ 
$25$  & $43.68_{\pm3.40}$   & $52.69_{\pm4.87}$ & $68.80_{\pm6.76}$ &  $75.33_{\pm7.38}$ & $64.04_{\pm0.79}$ & $76.85_{\pm0.94}$ & $83.71_{\pm0.41}$ & $65.65_{\pm5.85}$ & $81.79_{\pm4.36}$ & $87.84_{\pm2.38}$ \\ 
$50$  & $29.70_{\pm0.37}$   & $45.18_{\pm2.91}$ & $62.23_{\pm6.10}$ &  $67.63_{\pm5.78}$ & $54.17_{\pm0.90}$ & $70.01_{\pm0.84}$ & $76.79_{\pm0.43}$ & $56.67_{\pm5.30}$ & $73.80_{\pm3.18}$ & $81.55_{\pm0.05}$ \\ 
$100$ & $29.61_{\pm0.35}$   & $18.50_{\pm1.83}$ & $40.15_{\pm1.17}$ & $44.52_{\pm1.74}$ & $24.01_{\pm5.08}$ & $55.70_{\pm1.17}$  & $63.31_{\pm1.25}$ & $55.43_{\pm1.09}$ & $72.73_{\pm0.31}$ & $79.78_{\pm1.08}$ \\
\bottomrule 
\end{tabular}
}
\label{table: scalibility_booksum}
\end{table*}

\begin{table}[ht]
\caption{
Source attribution accuracy for $500$ data providers on the Reddit Webis-TLDR-17 dataset. 
}
\centering
\begin{tabular}{cc|ccc}
\toprule
n & method & acc.  & top-$3$. & top-$5$. \\
\midrule
\multirow{2}{*}{500}  & BM25   & 19.02 & 30.52  &  36.01  \\ 
  & WASA   & \textbf{35.66} & \textbf{48.65}  &  \textbf{54.39} \\ 
\bottomrule 
\end{tabular}
\label{table: reddit acc}
\end{table}

\subsection{Transferability}
\label{other key properties}
Our generated watermarked text has \emph{the same structure} as the watermarked text used to train our \algllm: They both embed $10$-character watermarks into texts with characters from the same vocabulary.
So, our generated watermarked text can be readily used as training data for other LLMs that, like our \algllm, can also generate synthetic text with watermarks.
That is, our generated watermarked text is \textbf{transferable} to other LLMs as their training data.

\subsection{Adaptability}
\label{app:adaptability}
Our \alg~framework only requires mild modifications to existing LLMs (Sec.~\ref{training watermark}) and can hence be easily adapted to fit various LLMs.
This has been empirically verified by our results in Secs.~\ref{subsec:exp:accuracy}\&\ref{scalability} and App.~\ref{app:more:experiments:acc:provenance}\&\ref{app:more:exp:scalability} that given the same experimental setup, accurate source attributions can be achieved by \algllm~that is obtained from our second-stage pre-training of various LLMs (i.e., GPT2, OPT, Llama2). 


\section{Detailed Results from Ablation Studies}
\label{app:more:ablation}
Here, we will present detailed results from our ablation studies.
In all our ablation studies, we use second-stage pre-training of the GPT2-Large model on the ArXiv dataset to obtain \algllm. 

\subsection{Effectiveness of our \algllm~Training}
\label{app:ablation:more:results:effectiveness}

We have mainly implemented two important algorithmic designs to help our \algllm~learn an accurate texts-to-watermarks mapping (Sec.~\ref{training watermark}): (1) using a designated embedding space for watermark tokens and (2) separating the prediction/generation spaces for the word and watermark tokens.
Here, we compare our \algllm~with two baselines: 
\emph{tokenizerGPT} implementing only the first design of a designated embedding space for watermark tokens, and \emph{originalGPT} (original GPT2-Large) implementing neither design.
We apply our second-stage pre-training to both baselines using the same (watermarked) data from the ArXiv dataset which was used for second-stage pre-training of the GPT2-Large model to obtain our \algllm, and evaluate the source attribution accuracy following that of Sec.~\ref{subsec:exp:accuracy}.
The results in Tab.~\ref{table: baseline_watermark} show that 
the first design alone does not improve the source attribution accuracy whereas the combination of both designs brings about a significant improvement. 
This is because merely creating the embedding space for watermark tokens does not help in learning the mapping from the texts to watermarks, and it is of particular importance to combine both designs for our \algllm~to perform well.
Moreover, our \algllm~achieves a significantly better source attribution accuracy at the expense of incurring 
more computational time. Note that \emph{originalGPT} takes longer training time than \emph{tokenizerGPT} because there is no designated embedding space for watermark tokens in \emph{originalGPT}, hence resulting in more training instances used.

\begin{table}[h]
    \caption{
    Comparison of source attribution accuracy achieved by \algllm~(obtained from second-stage pre-training of the GPT2 model) vs.~the baseline models on the ArXiv dataset where `n\_wtm' denotes the number of generated sentences with watermark, and `acc.' denotes the source attribution accuracy. 
    }
    \centering
    \begin{tabular}{l|cccc}
    \toprule
    model & n\_wtm & acc.  & n\_samples & training time \\ 
    \midrule
    BM25  & - & $54.73$  &-& - \\
    originalGPT  & $412$ & $45.69$ & $163507$ &  $6$h$30$m$3$s  \\ 
    tokenizerGPT & $439$ & $44.01$ & $140599$ &  $5$h$3$m$6$s  \\ 
    \algllm & $448$ & $74.84$ & $159387$ &   $8$h$9$m$24$s    \\ 
    \bottomrule
    \end{tabular}
    \label{table: baseline_watermark}
\end{table}

\subsection{Strategy for Selecting Sentences to Watermark}
\label{sec:random}
As we have discussed in Sec.~\ref{embed_watermark}, for every data provider, we embed watermarks into the sentences with top TF-IDF scores and then use these watermarked sentences 
for the second-stage pre-training (Sec.~\ref{training watermark}) of the GPT2 model to obtain our \algllm.
This is because the sentences with high TF-IDF scores are more representative of the text data from a data provider, which makes it easier to learn the mapping from the texts of different data providers to their corresponding unique watermarks.
Here, we will evaluate whether this strategy is effective by comparing it with the natural baseline of randomly selecting sentences to embed watermarks.
The results in Tab.~\ref{table: embedding} show that when selecting $20\%$ of the sentences for watermarking, the strategy of random embedding decreases the source attribution accuracy, which validates the effectiveness of our strategy of selecting sentences with high TF-IDF scores to watermark.

\begin{table}[h]
\caption{
Source attribution accuracy 
achieved by \algllm~(obtained from
second-stage pre-training of the GPT2 model on the ArXiv dataset) 
using different strategies to select the sentences for watermarking.
}
\centering
\begin{tabular}{l|ccc}
\toprule
embedding strategy & acc. & top-$3$. \\
\midrule
TF-IDF (ours)  & $74.84$  &  $95.76$ \\ 
Randomly Embed  & $71.40$  &  $94.48$ \\
\bottomrule 
\end{tabular}
\label{table: embedding}
\hspace{1.5mm}
\end{table}

\begin{table}
\caption{Comparison of source attribution accuracy and perplexity
achieved by \algllm~(obtained from
second-stage pre-training of the GPT2 model on the ArXiv dataset)
across different dataset sizes.}
\centering
\begin{tabular}{l|cccc}
\toprule
dataset size  & acc. &  top-$3$.& perplexity \\
\midrule
$10\%$: $100$MB   &  $68.80$   &  $94.10$ &  $14.6135$ \\ 
$33\%$: $300$MB   &  $74.84$   &  $95.76$ &  $12.6570$  \\ 
$66\%$: $600$MB   &  $76.28$   &  $95.88$ &  $11.6749$ \\ 
$100\%$: $1$GB    &  $78.48$   &  $95.80$ &  $11.3171$ \\ 
\bottomrule
\end{tabular}
\label{table: dataset_size}
\end{table}

\subsection{Impact of Enforced Watermark Generation}
\label{sec:force}

As discussed in Sec.~\ref{subsec:exp:accuracy},  to evaluate the source attribution accuracy in our experiments, we have adopted a simple technique to enforce watermark generation in order to simplify the evaluations.
That is, if a watermark is not generated after the generation of the sentence is completed, we add the token $[WTM]$ to the end of the sentence to enforce the watermark generation.
Here, we will evaluate the impact of this enforced watermark generation.
The results in Tab.~\ref{table:force} show that the forcefully generated watermarks and naturally generated watermarks have comparable source attribution accuracy.
This shows that the technique of enforced watermark generation we have adopted has minimal impact on the evaluations of the source attribution accuracy (Sec.~\ref{subsec:exp:accuracy}).


\begin{table*}[htb]
\centering
\caption{Source attribution 
accuracy 
achieved by \algllm~(i.e., obtained from second-stage pre-training of the GPT2 model on the ArXiv dataset)
for naturally generated watermarks (denoted by `watermark\_nf') vs.~forcefully generated watermarks (denoted by `watermark\_f').
}
\resizebox{0.8\linewidth}{!}{
\begin{tabular}{l|ccc|ccc}
\toprule
category  &  n\_watermark\_nf & n\_match\_nf  & acc.\_nf  & n\_watermark\_f & n\_match\_f & acc.\_f\\ 
\midrule
hep-th    & $45.8$ & $30.4$  &  $66.38$   &$4.2$& $2.4$ & $57.14$         \\ 
hep-ph    & $44.2$ & $37.8$  &  $85.52$   &$5.8$& $4.8$ & $82.76$      \\ 
quant-ph  & $46.0$ & $35.4$  &  $77.00$   &$4$& $2$ &5$0.00$              \\ 
astro-ph  & $44.2$ & $38.6$  &  $87.33$  & $5.8$ & $4.6$ &$79.31$        \\ 
cs.CV     & $44.2$ & $36.4$ &  $82.35$   & $5.8$ &  $4.6$  &$79.31$       \\ 
cs.LG     & $44.4$ & $35.0$ &  $78.83$    & $5.6$ &  $3.8$  &$67.86$         \\ 
cond-mat.mes-hall   & $44.8$ & $28.8$ &  $64.29$ & $5.2$ & $3.6$ & $69.23$            \\ 
gr-qc     & $43.2$ & $33.8$  &   $78.24$    &$6.8$& $4.4$  & $64.71$      \\ 
cond-mat.mtrl-sci  & $46.6$ & $30.6$&   $65.67$    &$3.4$& $1.8$  &$52.94 $    \\ 
cond-mat.str-el  &$ 44.6$ & $31.6$  & $70.85$     &$5.4$&  $3.8$  &$70.37$   \\ 
\midrule
Total   & $448$ & $338.4$ &  $75.54$    &$52$& $35.8$ & $68.85$  \\ 
\bottomrule 
\end{tabular}
}
\label{table:force}
\end{table*}

\subsection{Unattributable Content Analaysis}
\label{app:unattributable}
Here we consider the special case where the LLM-generated content is not attributable to any data provider. Note that in our main experiments, such a case does not exist since all data providers have watermarked their training data. Such unattributable content might be generated from public datasets used for pretraining the LLM, but we do not consider attributing sources to the public datasets in this paper as stated in Sec.~\ref{characteristics}; instead, we have focused on attributing to the data providers' watermarked private datasets. Moreover, it is hard to design prompts to enforce the model to generate content only from pretrain-knowledge, making it difficult to design corresponding experiments. Therefore, here we choose the setting by training our framework with data from both $5$ watermarked data providers and $5$ unwatermarked data providers to force our \algllm~to be able to generate content that is not attributable to the watermarked data providers. In this setting, our framework generates watermarks for $12\%$ of the sentences generated from the $5$ unwatermarked data providers while generating watermarks for $87.6\%$ of the sentences generated from the $5$ watermarked data providers. By analyzing the watermarks for sentences from unwatermarked data providers, we observe that $100\%$ of these watermarks are from the watermarked data providers. This suggests that if there exists content not attributable to any data provider, our framework sometimes might misclassify it to the watermarked data providers.

\subsection{Effectiveness of \alg~for Supervised Finetuning (SFT) Task}
\label{app:SFT:task}
In this section, we show that our \alg~framework can be effective for SFT tasks as well. Overall, while finetuning for the SFT task, our \algllm~can also learn the mapping from the texts of the data providers to their unique watermarks using an algorithm akin to the one described in Sec.~\ref{training watermark}.
Then, during sample prediction, our \algllm~can provide not only the predicted label but also the corresponding watermark.

\begin{figure}[h]
    \centering 
    \includegraphics[width=0.7\linewidth]{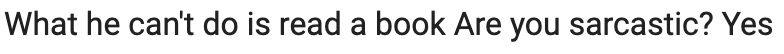} 
    \caption{Example of training samples in the SFT dataset.}
    \label{fig: sft example}
\end{figure}

Specifically, for the SFT task, we apply prompt finetuning~\citep{gao-etal-2021-making} where we introduce a prompt (manual template) after each training data. We then introduce the watermark following the training data by embedding it after the label.
Each supervised data point $s_i$ is a sequence of tokens: $s_i = [u_1, u_2, \ldots, u_{|s_i|}]$ where $|s_i|$ is the token count for $s_i$. For instance, $s_i =$ ``What he can't do is read a book'' in Fig.~\ref{fig: sft example}. We extend $s_i$ by appending a template, which results in $s_i^{\text{template}} = [u_1, u_2, \ldots, u_{|s_i|}, u_{|s_i|+1}, \ldots, u_{|s_i|+p}]$  with the template example being ``Are you sarcastic? Yes/No''. A data point embedded with a watermark is denoted as ${{s_i}^{\text{template}}}' = [u_1, u_2, \ldots, u_{|s_i|+p}, w_1, \ldots, w_m]$ where $w$'s represent watermark tokens. As shown in Fig.~\ref{fig: sft example}, an invisible watermark may follow after the label ``Yes''.

The training objective of \algllm~for SFT is a combination of maximizing the probability of label
word prediction and the probability of watermark generation. Since we only need to predict the label word, the predictive distribution can be simplified to 
\begin{equation}
\begin{array}{l}
P(u_{|s_i|+p} | u_1, u_2, \ldots, u_{|s_i|}, u_{|s_i|+1}, \ldots, u_{|s_i|+p -1}) \vspace{1mm}\\
= h_l[|s_i|+p-1] \cdot W_e^{\top}[\text{label word indices}]
\end{array}
\end{equation}
where $W_e^{\top}[\text{label word indices}]$ means to only use the label words' embedding. So, 
\begin{equation*}
\begin{array}{c}
  \displaystyle L_{\text{sft}}({s_i^{\text{template}}}') =\log P_u(u_{|s_i|+p} | u_1, u_2, \ldots, u_{|s_i|+p -1})\ ,
\end{array}
\label{eq:separate:likelihood:word_sft}
\end{equation*}
\begin{equation*}
\begin{array}{l}
L_{\text{wtm}}({s_i^{\text{template}}}') \vspace{1mm}\\
= \textstyle\sum_{j=1}^{m} \log P_w(w_j|u_1, u_2, \ldots, u_{|s_i|+p}, w_1, \ldots, w_{j-1})\ .
\end{array}
\label{eq:separate:likelihood:water_sft}
\end{equation*}
Then, the loss involves a combination of loss for label prediction, specifically in predicting the label word (i.e., Yes/No in the case of sarcasm), and loss for watermark generation. In particular, the loss is $Loss_{\text{\algllm}}({s_i^{\text{template}}}') = Loss_{\text{sft}}({s_i^{\text{template}}}') +  Loss_{\text{wtm}}({s_i^{\text{template}}}')$ in which
\begin{equation*}
\begin{array}{c}
\displaystyle
Loss_{\text{sft}}({s_i^{\text{template}}}') = \text{CE}(P(u_{|s_i|+p}), u_{|s_i|+p})\ ,\\ 
Loss_{\text{wtm}}({s_i^{\text{template}}}') = \sum_{j=1}^{m} \text{CE}(P_w(w_{j}), w_{j})\ .
\end{array}
\label{eq:separate:loss:word_sft}
\end{equation*}
To demonstrate the effectiveness of \algllm~for SFT data, we conduct experiments using the Self-Annotated Reddit Corpus (SARC)~\citep{khodak-etal-2018-large} which is an SFT dataset. This dataset, which is designed for sarcasm detection, includes $1.3$ million sarcastic comments sourced from Reddit; Tab.~\ref{table: reddit_sft} shows the details of this dataset. The dataset contains a column named `subreddit' which indicates the sub-forums dedicated to specific topics. Different subreddits are used to represent various data providers. Similar to the setting in Sec.~\ref{sec:property_val}, we select $10$ data providers in the experiment. We calculate the TF-IDF scores of all training points from each data provider and select those with the top $50\%$ of the TF-IDF scores (i.e., most representative sentences) for watermarking. We also adopt GPT2-Large as the pre-trained model. For the sarcasm task's template, we adopt the Question Prompt~\citep{liu-etal-2023-prompt}. Then, in terms of evaluating the source attribution accuracy, we randomly select each data point as the input/prompt to the trained \algllm~and use the subreddit of that data point as the source. The other evaluation settings are the same as that in Sec.~\ref{subsec:exp:accuracy}.

Tab.~\ref{table: sft result} illustrates that a top-$1$ source attribution accuracy of $50.80\%$ and a top-$3$ accuracy of $78.80\%$ can be achieved using our~\algllm. The performance is inferior compared to that observed in generation tasks, primarily due to the increased challenge in learning mappings from texts to watermarks because texts in the SFT dataset contain fewer tokens on average. Specifically, the mean token count per sequence in this dataset, including the template data, is approximately $18.4$  which contrasts with the average of $512$ tokens per sequence in unsupervised tasks. Despite this, the achieved accuracy significantly surpasses the baseline of $10.00\%$.
Furthermore, the model exhibits a decent sarcasm prediction accuracy of $86.60\%$ which even surpasses the performance of the original GPT2. One of the reasons may be that certain subreddits are more likely to contain sarcastic comments and our watermarking framework coincidentally captures this pattern. The results demonstrate that our \alg~framework is still effective for SFT data and can maintain the performance preservation property.

\begin{table}[h]
    \caption{Comparison of performances of the original GPT2 model trained with unwatermarked data and our \algllm~in terms of sarcasm prediction accuracy (`pred acc') and source attribution accuracy (`acc' and `top-3').}
    \centering
    \begin{tabular}{l|cccc}
    \toprule
    model & pred acc.  & acc. & top-3. & training time \\ 
    \midrule
    random  & $50.00$ & $10.00$ & $30.00$ & - \\
    unwatermarked  & $84.80$ & - & - &  $3$h$37$m$38$s \\ 
    \algllm & $86.60$ & $50.80$  & $78.80$ &  $4$h$32$m$17$s  \\ 
    \bottomrule
    \end{tabular}
    \label{table: sft result}
\end{table}

\begin{table}[ht]
\caption{Information on the Self-Annotated Reddit Corpus (SARC) dataset.}
    \centering
    \begin{tabular}{p{4cm}|cc}
    \toprule
        & Training & Evaluation   \\    
    \midrule
    Comments                   &   $910$K    &  $101$K     \\ 
    Unique tokens            &   $464$K   &   $109$K    \\ 
    Total tokens               &   $9.5$M    &   $1$M   \\ 
    \bottomrule
    \end{tabular}
\label{table: reddit_sft}
\end{table}

\subsection{Relative Positions of Generated Watermarks}
\label{app:ablation:pos:of:wtm}
\begin{figure}[ht]
\centering
     \begin{tabular}{c}
         \includegraphics[width=0.55\linewidth]{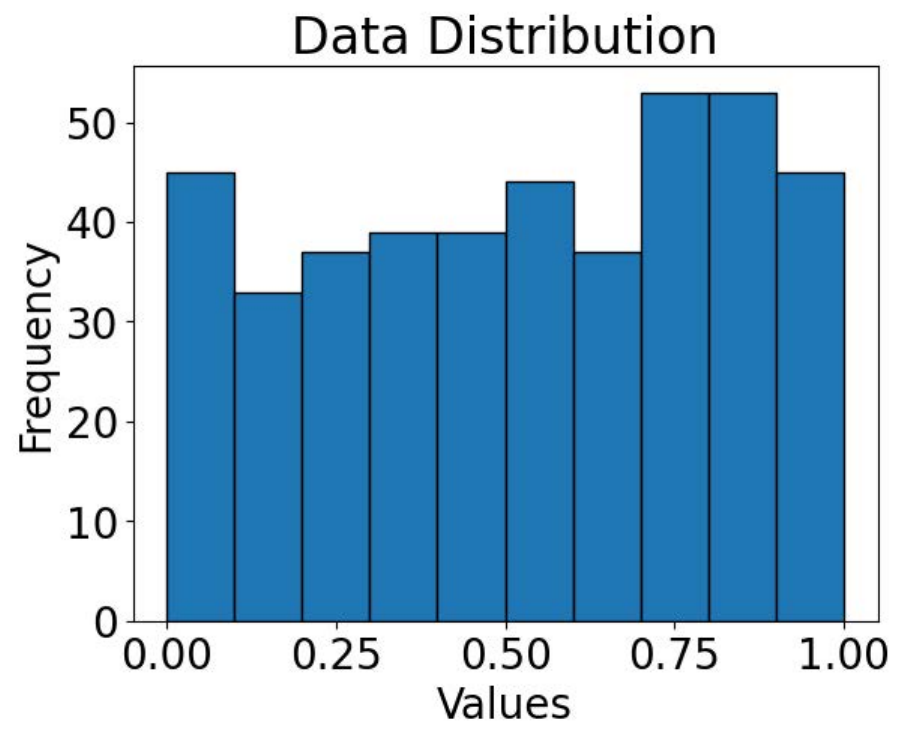}
     \end{tabular}
    \caption{Distribution of the relative positions of the generated watermarks in the generated sentence.
    }
    \label{fig: pos}
\end{figure}

To further investigate the nature of our generated watermarks, we have analyzed the distribution of the relative positions of the generated watermarks in the generated sentences. As shown in Fig.~\ref{fig: pos}, the generated watermarks are uniformly distributed within a sentence. This is because when we embed watermarks into the selected sentences for LLM training, the position of the embedded watermark is randomly selected. Therefore, after the LLM is trained, the position of the generated watermark in the generated sentence is also uniformly distributed. This uniform distribution of watermarks makes it harder for an adversary to remove the watermark, compared to the scenario where the watermarks are at a fixed position.

\subsection{Impact of Number of Watermarks in Training Data }
\label{app:ablation:more:results:perc:wtm}

\begin{figure}[h]
    \centering \includegraphics[width=0.49\linewidth]{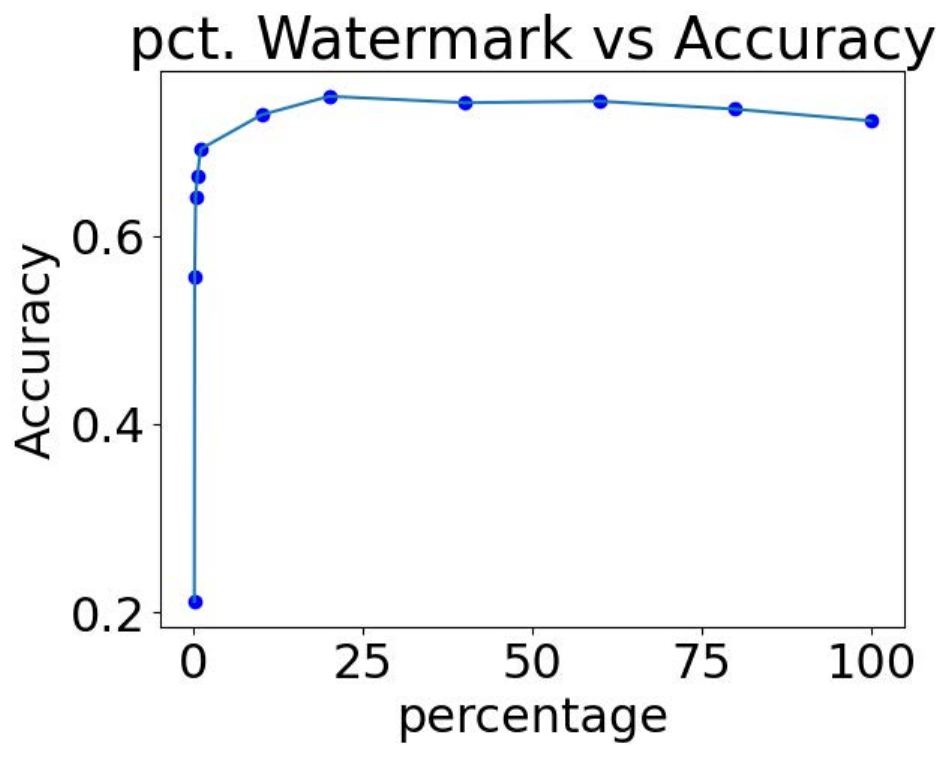}
    \centering \includegraphics[width=0.49\linewidth]{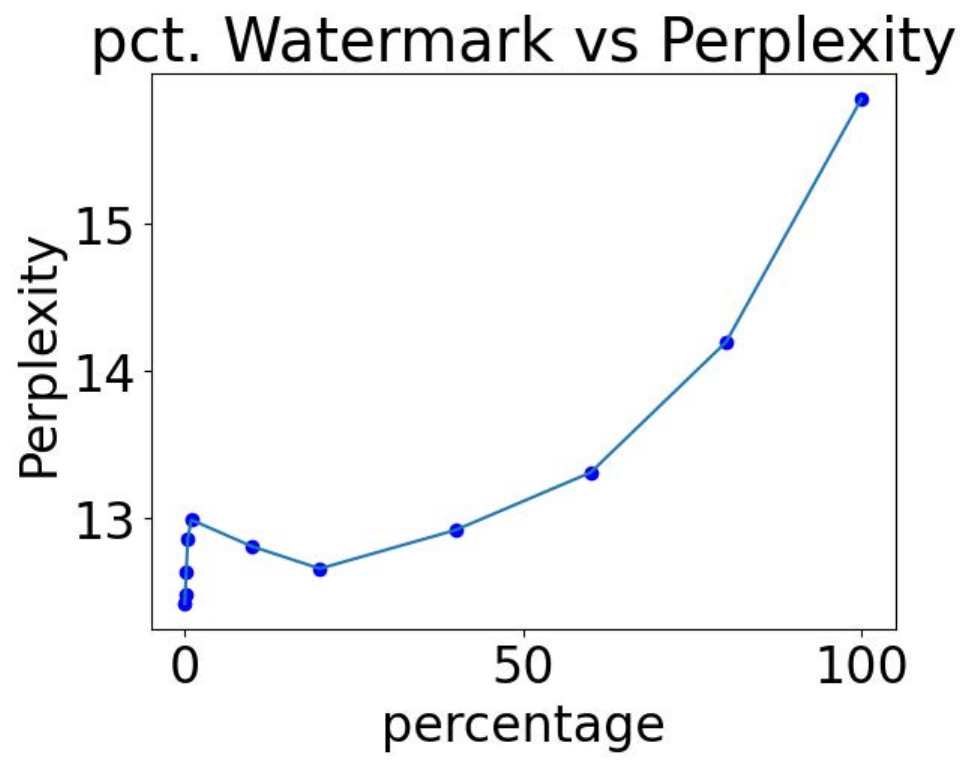}
    \caption{Source attribution accuracy and perplexity 
    achieved by \algllm~(i.e., obtained from second-stage pre-training of the GPT2 model on the ArXiv dataset) 
    vs.~percentage of watermarked sentences in the training data.}
    \label{fig:watermark_L}
\end{figure}
\begin{figure}[ht]
    \centering \includegraphics[width=0.49\linewidth]{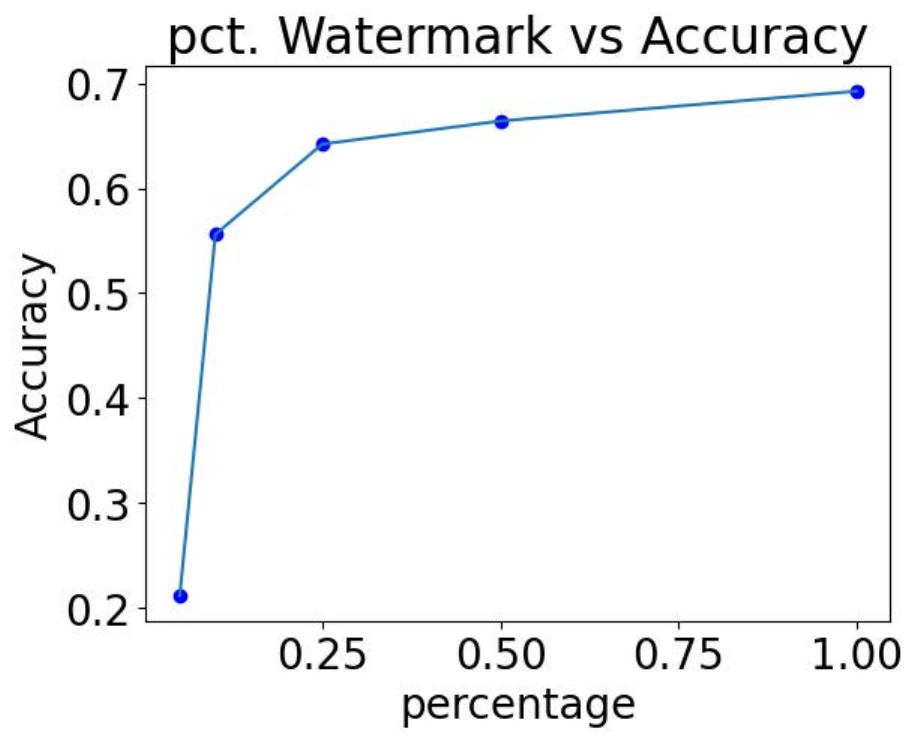}
    \centering \includegraphics[width=0.49\linewidth]{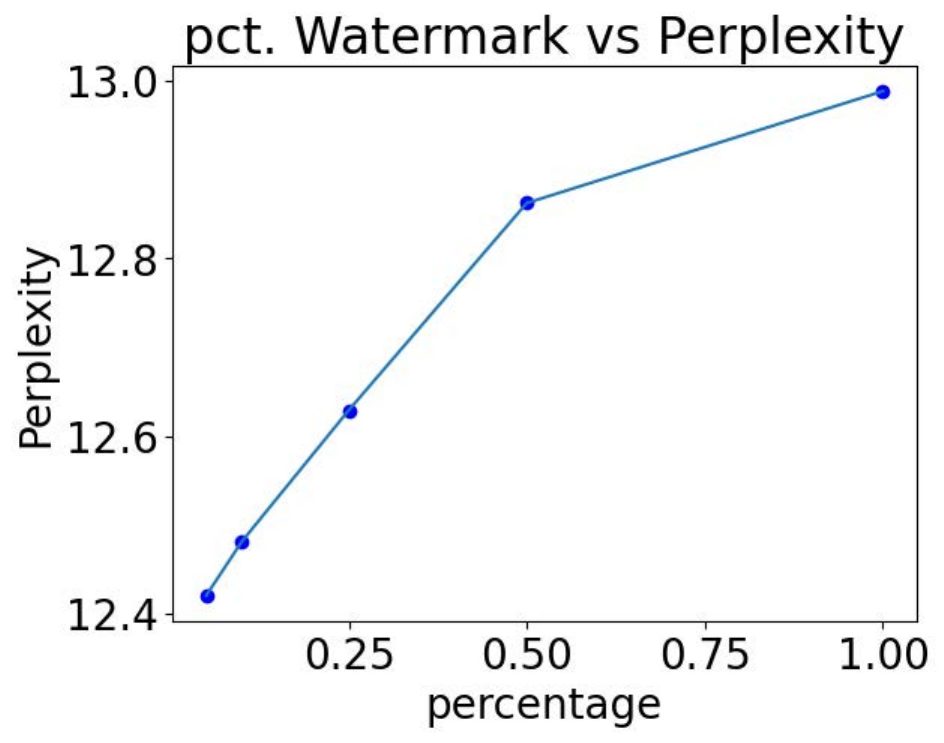}
    \caption{Source attribution accuracy and perplexity achieved by \algllm~(i.e., obtained from second-stage pre-training of the GPT2 model on the ArXiv dataset) vs.~percentage of watermarked sentences in the training data on a smaller scale of $0.05\%-1\%$ for a clearer visualization.}
    \label{fig:watermark_S}
\end{figure}

Here, we will evaluate the impact of the number of watermarks in the training data on the source attribution accuracy achieved by \algllm.
Following that of Sec.~\ref{embed_watermark}, we vary the percentage of sentences 
selected for watermarking (i.e., top $X\%$ of the TF-IDF scores) and evaluate its impact on our \algllm~obtained from
second-stage pre-training of the GPT2 model on the ArXiv dataset.
Fig.~\ref{fig:watermark_L} (left) shows that as the number of watermarks increases, the source attribution accuracy firstly increases and then declines.
This is because an overly small number of watermarks results in insufficient data for learning an accurate texts-to-watermarks mapping; meanwhile, if watermarks are added to an excessively large number of sentences, then some of the watermarked sentences \emph{may not be representative of the texts from their data providers}, which also increases the difficulty of learning the mapping from the texts of the data providers to their unique watermarks (see Sec.~\ref{embed_watermark}).
In addition, Fig.~\ref{fig:watermark_L} (right) shows that increasing the number of added watermarks in general leads to worse text generation performances (i.e., larger perplexity) of the \algllm.
The detailed results are provided in Tab.~\ref{table:percentage}. Moreover, Fig.~\ref{fig:watermark_S} shows a clearer visualization of the results in smaller percentages.

\begin{table}[h]
\caption{Comparison of source attribution accuracy achieved by \algllm~(i.e., obtained from second-stage pre-training of the GPT2 model on the ArXiv dataset) for different percentages of watermarked sentences in the training data.
The percentage of blocks that are watermarked is given as well.}
\centering
\begin{tabular}{ll|cccc}
\toprule
pct. sentences & pct. blocks & acc. & top-$3$. & perplexity \\ 
\midrule
$20\%$   & $88.25\%$ &  $74.84$  & $95.76$ & $12.6570$ \\ 
$40\%$   & $96.88\%$ &  $74.16$  & $95.45$ & $12.9180$ \\ 
$60\%$   & $98.86\%$ &  $74.32$  & $95.04$ & $13.3096$  \\ 
$80\%$   & $99.38\%$ &  $73.48$  & $95.40$ & $14.1952$  \\ 
$100\%$  & $100.00\%$ & $72.24$  & $95.00$ & $15.8465$  \\ 
\bottomrule
\end{tabular}
\label{table:percentage}
\end{table}

\subsection{Impact of Lengths of Conditioned Sentence and Generated Sentence}
\label{sec:length}
Recall that in our main experiments, we have used a sentence with $200$ characters as the input/prompt (i.e., the conditioned sentence) to our  \algllm, and let the \algllm~generate synthetic texts with $100$ tokens (Sec.~\ref{subsec:exp:accuracy}).
In this section, we vary the character lengths of both the conditioned sentence and the generated synthetic texts, and evaluate their impact on the source attribution accuracy achieved by \algllm~(i.e., obtained from
second-stage pre-training of the GPT2 model on the ArXiv dataset).
The results in Tab.~\ref{table: sentence_length} show that longer conditioned sentences (i.e., inputs/prompts) lead to better performances.
Moreover, when the length of the conditioned sentences is fixed (at $200$), increasing the length of the generated synthetic texts consistently reduces the number of forcefully generated watermarks (App.~\ref{sec:force}) while preserving the source attribution accuracy achieved by \algllm.


\begin{table}[h]
\caption{
Impact of the lengths of the conditioned sentences (inputs/prompts) and the generated synthetic sentences on the source attribution accuracy achieved by \algllm~(obtained from
second-stage pre-training of the GPT2 model on the ArXiv dataset) where 
`len.~cond.' stands for the character length of the conditioned sentences, `tokens syn.' refers to the number of tokens in the generated synthetic sentences, and `pct.~wtm\_f' denotes the percentage of forcefully generated watermarks.
}
\centering
\begin{tabular}{cc|ccc}
\toprule
len.~cond.  & tokens syn. & acc.  &  top-$3$. & pct.~wtm\_f \\
\midrule
$100$    &  $100$   &  $63.92$ & $89.96$ &  $15.2\%$  \\ 
$100$    &  $200$   &  $64.36$ & $89.48$ &  $5.2\%$  \\
$200$    &  $100$   &  $74.84$ & $95.76$ &  $8.6\%$   \\ 
$200$    &  $200$   &  $75.20$ & $95.64$ &  $4.2\%$   \\ 
$200$    &  $300$   &  $74.24$ & $95.40$ &  $2.2\%$   \\ 
$200$    &  $400$   &  $74.60$ & $95.24$ &  $1.0\%$  \\ 
\bottomrule
\end{tabular}
\label{table: sentence_length}
\end{table}


\subsection{Impact of Length of Watermark}
\label{app:ablation:len:of:wtm}
In our main experiments, we have adopted a watermark design that consists of $10$ characters/tokens (Sec.~\ref{embed_watermark}).
However, our \alg~framework allows for the use of watermarks with different lengths.
Here, we will test the impact of the length of the watermarks on the source attribution accuracy achieved by \algllm~(obtained from
second-stage pre-training of the GPT2 model on the ArXiv dataset).
The results in Tab.~\ref{table: length} show that for watermarks with $5$, $10$, and $15$ characters, their source attribution accuracies are comparable while the $5$-character watermark achieves slightly better performances.
This is likely because when the watermark is shorter, the resulting watermark prediction problem becomes relatively easier (i.e., the number of parameters in the last linear layer is smaller), which may lead to better watermark prediction and generation.
However, note that a long watermark is favored when there is a need to scale to a large number of data providers.
Therefore, our \alg~framework offers the flexibility to choose watermarks with different lengths, and the preferred watermark length can be application-dependent.


\begin{table}[ht]
\caption{
Source attribution accuracy achieved by \algllm~(obtained from
second-stage pre-training of the GPT2 model on the ArXiv dataset) using watermarks with different lengths.
}
\centering
\begin{tabular}{l|ccc}
\toprule
len.~watermarks & acc. & top-$3$.  \\
\midrule
$5$ \space\space   characters &   $76.12$   & $95.48$ \\ 
$10$ characters &   $74.84$   &  $95.76$ \\
$15$ characters &   $74.12$   & $95.28$  \\
\bottomrule 
\end{tabular}
\label{table: length}
\end{table}

\begin{table}
\caption{Source attribution accuracy achieved by \algllm~(obtained from
second-stage pre-training of the GPT2 model on the ArXiv dataset) after training with more epochs.}
\centering
\begin{tabular}{c|ccc}
\toprule
n\_epochs  & acc. & top-$3$. \\
\midrule
$1$ & $74.84$   &  $95.76$   \\ 
$2$ & $76.96$   &  $96.00$   \\
$3$ & $75.88$   &  $95.88$   \\
\bottomrule 
\end{tabular}
\label{table:overfit}
\end{table}

\subsection{Impact of Number of Watermark Characters}
\label{app:ablation:num:of:char}
In our main experiments, we have used $6$ invisible Unicode characters to form each character in the $10$-character watermark. Our \alg~framework also allows for the use of watermarks such that each character in the watermark can be chosen among a different number of available characters. Tab.~\ref{table: available_watermarks} shows the source attribution accuracy achieved by \algllm~(obtained from
second-stage pre-training of the GPT2 model on the ArXiv dataset) when each character in the watermark can be chosen among only $2$ available characters: U+200B: Zero Width Space and
U+200C: Zero Width NonJoiner. The results are comparable while the one with $2$ available characters shows slightly worse top-$3$ accuracy. This is likely because when fewer available characters are used, the watermarks for different categories are more similar to each other, which may make top-$3$ classification more difficult.

\begin{table}[ht]
\caption{
Impact of the number of available characters (used to make up each character in the $10$-character watermark) on the source attribution accuracy achieved by \algllm~(obtained from
second-stage pre-training of the GPT2 model on the ArXiv dataset).  
}
\centering
\begin{tabular}{c|cc}
\toprule
n\_available\_characters  &  acc.  &  top-$3$.  \\
\midrule
$2$   &  $75.48$ & $89.92$   \\ 
$6$   &  $74.84$ & $95.76$  \\
\bottomrule
\end{tabular}
\label{table: available_watermarks}
\end{table}

\subsection{Impact of Amount of Data for Second-Stage Pre-training to Obtain \algllm}
\label{app:ablation:amount:of:data}
Here, we will evaluate the impact of using varying amounts of data from the ArXiv dataset for our second-stage pre-training (Sec.~\ref{training watermark}) of the GPT2 model to obtain \algllm.
As discussed in App.~\ref{subsec:exp:setup:datasets}, in our main experiments for the ArXiv dataset, we have used $33\%$ of text data from every category (i.e., data provider) to reduce computations.
Here, we will vary this percentage to evaluate its impact on both the source attribution accuracy and the text generation performance achieved by our \algllm.
The results in Tab.~\ref{table: dataset_size} demonstrate that as more data is used, both the source attribution accuracy and the text generation ability (i.e., perplexity) achieved by our  \algllm~are generally improved.


\subsection{Impact of Number of Training Epochs}
\label{sec:overfit}
As we have discussed in App.~\ref{subsec:exp:setup}, we have trained our \algllm~for one epoch during the second-stage pre-training (Sec.~\ref{training watermark}).
Here, we will evaluate the performance of \algllm~after training with more epochs. The results in Tab.~\ref{table:overfit} show that training with multiple epochs in general further improves the performance.
This demonstrates the potential of our \alg~framework to achieve even better source attribution accuracy (than those presented in our current experiments) with more computations.

\section{Case Studies}
\label{app:case_study}
\subsection{Generated Texts with Imperceptible Watermarks}
\label{app:generated_text}
We have discussed in Sec.~\ref{watermark generation} how our trained \algllm~can be used to generate synthetic texts with embedded watermarks.
Fig.~\ref{fig: synthetic} below shows an example of the watermarked texts generated by our \algllm, which verifies that the generated watermarks that are embedded into the generated texts are indeed imperceptible to human eyes. Therefore, the readability of the generated texts will not be affected much. 

\begin{figure}[htb]
    \centering
    \includegraphics[width=0.95\linewidth]{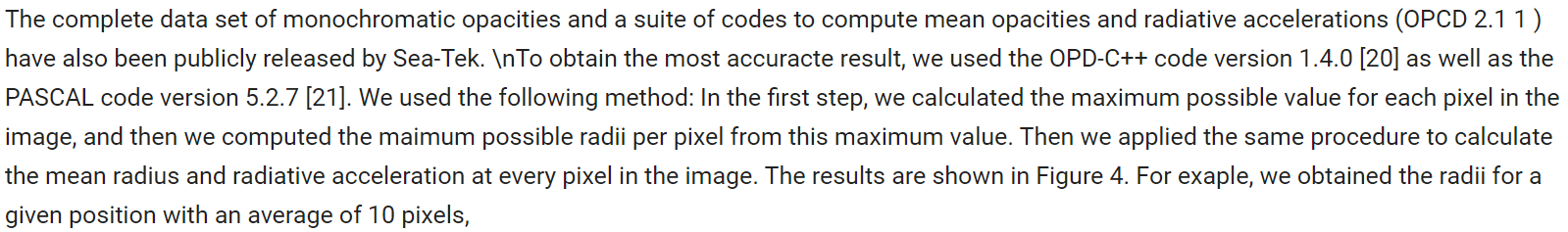}
    \caption{An example of our \algllm-generated synthetic texts with embedded watermarks that are imperceptible to human eyes.}
    \label{fig: synthetic}
\end{figure}

\subsection{Generated Data and its Source}

\begin{figure}
     \begin{tabular}{l}
         \includegraphics[width=0.95\linewidth]{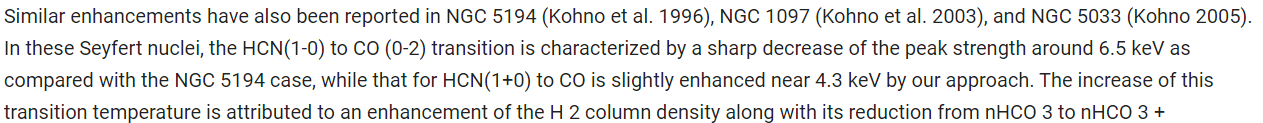}
     \end{tabular}
    \caption{Generated text from ArXiv dataset (\textit{astro-ph} category).}
    \label{fig: arxiv example}
\end{figure}

\begin{figure}
     \begin{tabular}{l}
         \includegraphics[width=0.95\linewidth]{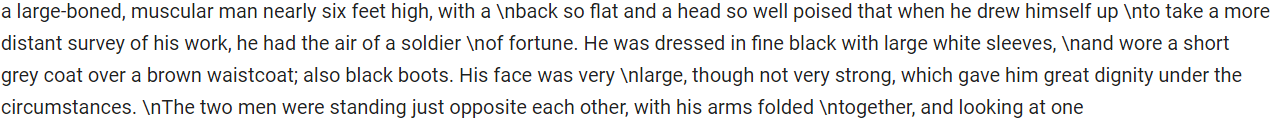}
     \end{tabular}
    \caption{Generated text from BookSum dataset (\textit{Adam Bede} category).}
    \label{fig: booksum example}
\end{figure}

To facilitate a better demonstration of the performance of our \alg~framework, we perform a case study on the synthetic data generated by our \algllm. The examples shown in Figs.~\ref{fig: arxiv example} and~\ref{fig: booksum example} are the generated texts from our \algllm~trained with the ArXiv dataset and the Booksum dataset, respectively. They further verify the invisibility of the generated watermarks and demonstrate that our framework preserves the quality of the generated texts.

\subsection{Generated Data with Two Source}
\label{case:multi:source}
Considering the special cases where the generated data is a combination of data from two providers, our current \alg~framework naturally handles them: We can use the generated top-$k$ watermarks to identify the $k$ most likely data providers in order to account for cases where there are multiple data providers.

\begin{figure}[ht]
     \begin{tabular}{c}
         \includegraphics[width=1\linewidth]{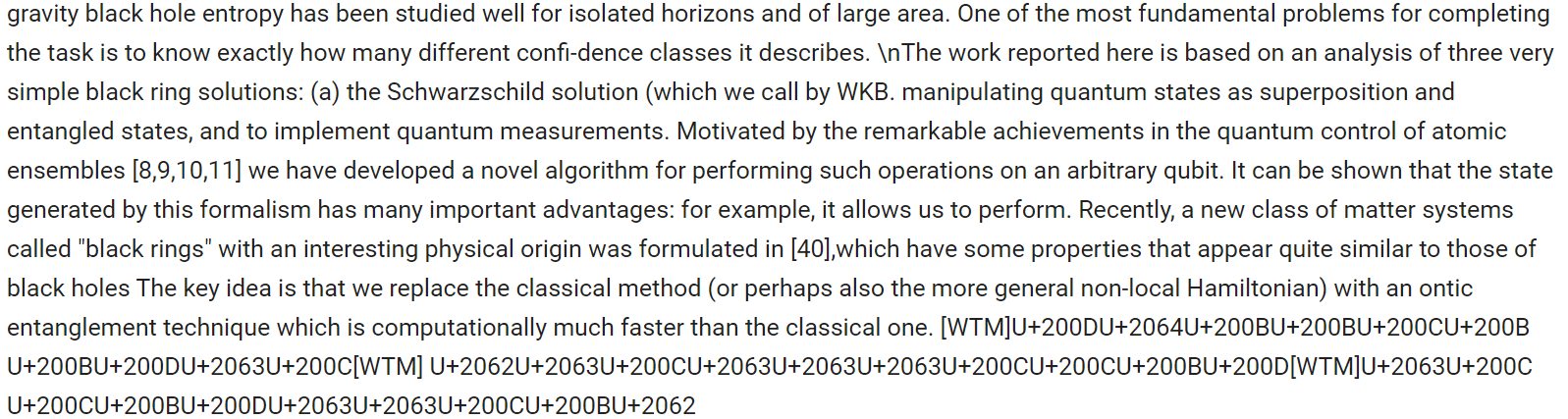}
     \end{tabular}
    \caption{Combined generated text from ArXiv dataset (\textit{gr-qc} and \textit{quant-ph} categories) with top-$3$ watermarking covering both watermarks.}
    \label{fig: multisource example 1}
\end{figure}

To demonstrate our framework's capability in this context, we have crafted several case studies simulating examples of text that are combinations of two data providers. We select two pieces of text generated by different data providers and manually concatenate them. Subsequently, we use the concatenated text as the prompt for\algllm~to generate the top-$3$ watermarks. As an example in Fig.~\ref{fig: multisource example 1}, we have crafted the texts as the concatenation of the generated texts from two data providers \textit{gr-qc} (with watermark `U+200DU+2064U+200BU+200BU+200CU+200 BU+200BU+200DU+2063U+200C') and \textit{quant-ph} (with watermark `U+2062U+2063U+200CU+2063U+2063U+20
63U+200CU+200CU+200BU+200D'). In such cases, our framework is able to produce the watermarks corresponding to both data providers among the top-$3$ generated watermarks. Note that in the above example and the next, we manually visualize the watermarks for illustrative purposes, while in real cases, the watermarks remain invisible.

\begin{figure}[ht]
     \begin{tabular}{c}
         \includegraphics[width=1\linewidth]{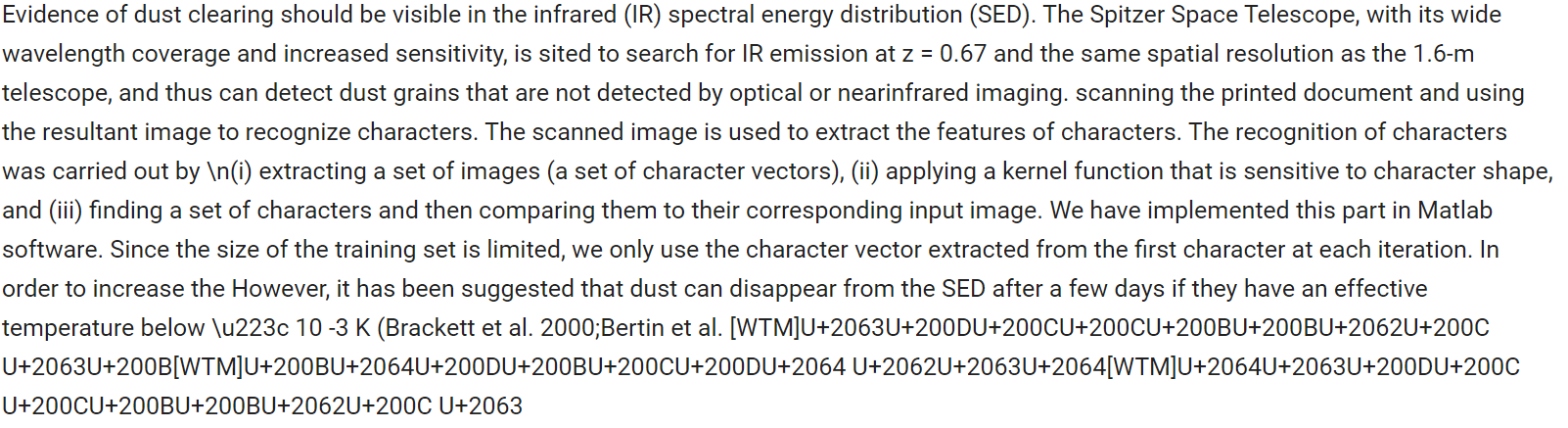}
     \end{tabular}
    \caption{Combined generated text from ArXiv dataset (\textit{astro-ph} and \textit{cs.CV} categories) with top-$3$ watermarking covering both watermarks.}
    \label{fig: multisource example 2}
\end{figure}

As another example, we have crafted the texts (i.e., shown in Fig.~\ref{fig: multisource example 2}) as the concatenation of the generated texts from another two data providers \textit{astro-ph} (with watermark `U+2063U+200DU+200CU+200CU+200BU+200B U+2062U+200CU+2063U+200B') and \textit{cs.CV} (with watermark `U+200BU+2064U+200DU+200BU+200CU+200D
U+2064U+2062U+2063 U+2064'). In this case, our framework is also able to generate the watermarks for both data providers among the top-$3$ watermarks. These results demonstrate the potential of our top-$k$ source attribution to handle scenarios in which the generated data is a combination of multiple data providers.

\section{Frequently Asked Questions}



\paragraph{The paper assumes data providers are willing to embed watermarks in their data to track usage, but in practice, they may prioritize data privacy over adding any extra information.}

Firstly, the objective of this work is to protect the IP rights of the data providers under the setting that there is a necessity to certify the source of online content produced by LLMs, as discussed in Sec.1. Under this setting, the data providers are willing to have their identity disclosed and attributed to. In practice, this setting may correspond to authors of academic papers who are willing to be identified and cited for their work.

Meanwhile, as discussed in App.~\ref{app:ethical:considerations}, in our WASA framework, only the watermark can be seen in the generated data, which does not imply personal information about the data providers. Therefore, data privacy can be preserved as long as the mapping from watermarks to data providers is kept confidential. In practice, if some data providers prioritize data privacy and do not want their identities to be revealed, they may request the LLM owner to not decode their watermarks and reveal them as sources to the public, in which case users will not be able to infer any private information from the watermark itself.

From another perspective, given our proposed watermarking scheme, data providers will also be able to check data provenance and see whether their watermarked data have been misused, which serves as a protection of data privacy in a different sense.

\paragraph{It seems the removal of all invisible characters could render the watermarks ineffective.}

Firstly, we have considered various scenarios where the generated watermark is modified or removed in our paper (Sec.~\ref{robustness} and App.~\ref{app:more:exp:robustness}). We have tested our watermark regeneration defense against these scenarios to regenerate the attacked watermark and preserve a high source attribution accuracy of $71.60\%$ (top-$3$ $93.76\%$), which is comparable to the original $74.84\%$ (top-$3$ $95.76\%$). Thus, \textit{our watermark regeneration is an effective defense mechanism} to address the straightforward removal of watermarks.

Secondly, we would like to consider the usage of our framework where source attribution is performed immediately as the LLM generates text together with the watermark. Under this setting, the identification of the data provider of the generated text takes place right after LLM generation and there would be no opportunity for attackers to modify the generated watermarks. In practice, this setting may correspond to the scenario that when the user queries an LLM, the source is provided along with the output of the LLM.

\paragraph{How does the evaluation, particularly the experimental setup correlate with realistic scenarios where LLMs generate novel content?}

In real-world scenarios, source attribution is more likely to be performed on LLM-generated content to find the source for the generation. In our evaluation, the source attribution accuracy is also measured on the generated sentence of the LLMs, using the sentences selected from the training datasets as inputs/prompts. Hence, our evaluation design aligns with the real-world source attribution applications on both performing on synthetic data. Note that we use the sentences from the training datasets as inputs/prompts to LLMs solely to decide the ground-truth source for the generated content: On the one hand, we can determine the source of the generated sentence directly as the source (training data provider) for the input/prompt (as validated in App. E.3); On the other hand, if we choose inputs/prompts as those we do not know the source, it would be more challenging to decide the source for the generated sentence and make the evaluation of source attribution less reliable.

Importantly, we have adopted various datasets in our experiments that correspond to different real-life use cases. The ArXiv and DBpedia datasets correspond to paper and knowledge attribution, while the BookSum dataset refers to story attribution. The CC-News, IMDB, and FakeNews datasets represent a more challenging use case: the attribution of word/expression usage.

\end{document}